%% file: arXiv.tex
\documentclass[10pt,twocolumn,letterpaper]{article}

\usepackage[pagebackref=true,breaklinks=true,colorlinks,bookmarks=false]{hyperref}

\usepackage[table,xcdraw]{xcolor}
\usepackage{iccv}
\usepackage{times}
\usepackage{epsfig}
\usepackage{graphicx}
\usepackage{amsmath}
\usepackage{breqn}
\usepackage{amssymb}
\usepackage{epstopdf}
\usepackage{booktabs}
\usepackage{color}
\usepackage{comment}
\usepackage{float}
\usepackage{svg}
\usepackage{mathtools}
\usepackage{wasysym}
\usepackage{multirow}

\usepackage{subfig}

\usepackage[toc,page]{appendix}

\iccvfinalcopy 

\ificcvfinal\fi

\begin{document}

\title{Learning Representations for Predicting Future Activities}

\author{Mohammadreza Zolfaghari\textsuperscript{1} \href{https://orcid.org/0000-0002-2973-4302}{\includegraphics[scale=0.3]{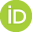}}, {\"O}zg{\"u}n {\c C}i{\c c}ek\textsuperscript{1} \href{https://orcid.org/0000-0002-8341-7332}{\includegraphics[scale=0.3]{images/orcid.png}}, Syed Mohsin Ali\textsuperscript{1}, Farzaneh Mahdisoltani\textsuperscript{2}, \\ Can Zhang\textsuperscript{3}, and
        Thomas Brox\textsuperscript{1} \href{https://orcid.org/0000-0002-6282-8861}{\includegraphics[scale=0.3]{images/orcid.png}} \\
\textsuperscript{1}University of Freiburg, \textsuperscript{2}University of Toronto,  \textsuperscript{3}Peking University\\
{\tt\small \textsuperscript{1}\{zolfagha,cicek,alis,brox\}@cs.uni-freiburg.de,} {\tt\small \textsuperscript{2}farzaneh@cs.toronto.edu,} {\tt\small \textsuperscript{3}zhangcan@pku.edu.cn}
}

\maketitle

\begin{abstract}
Foreseeing the future is one of the key factors of intelligence. It involves understanding of the past and current environment as well as decent experience of its possible dynamics. In this work, we address future prediction at the abstract level of activities. We propose a network module for learning embeddings of the environment's dynamics in a self-supervised way. To take the ambiguities and high variances in the future activities into account, we use a multi-hypotheses scheme that can represent multiple futures. We demonstrate the approach\footnote{Code: \url{https://github.com/lmb-freiburg/PreFAct}} by classifying future activities on the Epic-Kitchens and Breakfast datasets. Moreover, we generate captions that describe the future activities. 
\end{abstract}

\input{introduction.tex}
\input{relatedwork.tex}

\input{method.tex}
\input{multiHypo.tex}
\input{results.tex}

\input{conclusion.tex}

\begin{appendices}

\input{sup_mat.tex}
\end{appendices}

{\small
\bibliographystyle{ieee}

\input{arXiv.bbl}
}

\end{document}

%% file: introduction.tex
\section{Introduction}

The ability to foresee what possibly happens in the future is one of the factors that makes humans intelligent. Predicting the future state of the environment conditioned on the past and current states requires a good perception and understanding of the environment as well its dynamics. This ability allows humans to plan ahead and choose actions that shape the environment in our interest. 

In this paper, we focus on improving the model's understanding of the environment's dynamics by simply observing it. 
Due to the availability of unlabeled video data, self-supervised learning from observations is very attractive compared to approaches that require explicit labeling of large amounts of data.

    
    
    

\begin{figure}[t]
\centering
\includegraphics[trim=0mm 0mm 0mm 0mm,clip,width=0.48\textwidth]{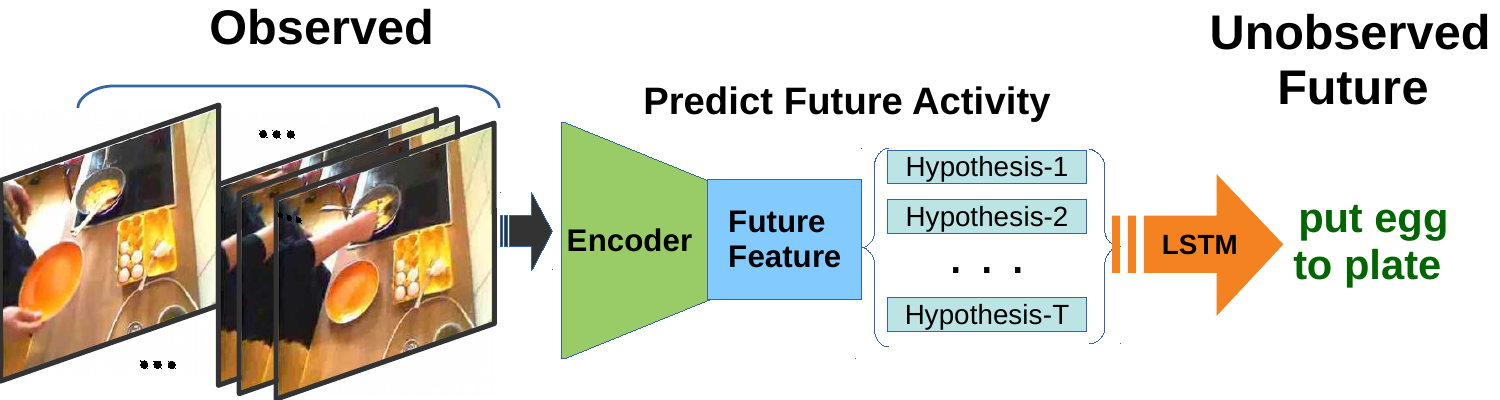}
\caption{ 
Future prediction at the activity level. From a spatio-temporal observation, we predict multiple feature hypotheses that suggest possible future activities. Each feature represents a future action and a future object. A language module generates a caption for this representation of the future. }
\label{teaser_fig}

\end{figure}

In comparison to literature on understanding the current state of the environment -- works typically known under the terms of \emph{semantic segmentation} or \emph{action classification} -- 
there is limited work addressing the problem of predicting future states. In this paper, we are interested in predicting future activities. 
Our work is different from most prior works on video prediction, as they focus on predicting whole frames of the future~\cite{liu2018ano_pred,vondrick15,Byeon_2018_ECCV}. In the context of decision making systems, pixel-wise future prediction is too detailed and cannot be expected to enable longer prediction horizons than just a few frames. The strategy of Luc et al.~\cite{paul_segmentation} to predict the segmentation of future frames by forecasting the future semantics instead of raw RGB values, appears much more promising. We follow a similar strategy to predict abstract features and even increase the level of abstraction by dealing with activities rather than pixel-wise labeling; see Figure \ref{teaser_fig}. 

This connects large part of the problem with activity classification: before making predictions about future states, we must interpret the given video input and extract features that describe the current state of the environment. In order to cover not only the present state but also the context of the past, we build on the work by Zolfaghari et al.~\cite{zolf18}. This work on activity classification samples frames from a large time span of the past and converts the context from these frames into a feature  representation optimized for classifying the observed activity. We argue that this feature representation is a good basis for learning a representation of what is likely to happen in the future. While we keep the first untouched, we learn the latter from the time-course of the videos. This even allows us to learn the dynamics in a self-supervised way. In this paper, we report results for both supervised (with action class labels) and self-supervised (unlabeled videos) training. 

The predicted future state is provided as an activity class label or as a caption generated by a captioning module based on the predicted representation. Since the future is non-deterministic, forcing the network to predict a single possible outcome leads to contradictory learning signals and will hamper learning good representations. Therefore, we use a multi-hypotheses scheme that can represent multiple futures for each video observation.

Moreover, we decouple the prediction of the action and the object involved in an activity. This allows the model to generalize the same action across multiple objects and learn from only few shots or even without observing all combinations during training. 

Overall, we propose the first approach for Predicting Future Activity (PreFAct) over large time horizons. Our method involves four important components: (1) a future prediction module that transforms abstract features of an observation to a representation of the future; (2) decoupling of the future representation into object and action; (3) representation of multiple hypotheses of the future; (4) natural language caption of the future representation. 

%% file: relatedwork.tex
\section{Related work}
\textbf{Future Image Prediction.} Many existing approaches for future prediction focus on generating future frames \cite{determ_1,determ_2,determ_3}. Since predicting RGB pixel intensities is difficult and the future is ambiguous, these methods usually end up in predicting a blurry average image. To cope with the non-determinism, Mathieu et al.~\cite{future_image1} suggest to use a multi-scale architecture with adversarial training. Stochastic approaches \cite{erhan_17,lee2018savp} use adversarial losses and latent variables to explicitly model the underlying ambiguities. 

However, pixel-level prediction is still limited to a few frames into the future especially when the scene is highly dynamic and visual cues change rapidly. Moreover, pixel-level fine-detailed future prediction is not necessary for many decision making systems. 
    
\textbf{ Future Semantic Prediction.} There are many works which also tackle future prediction in a more abstract way \cite{Soomro_2016_CVPR,pat_eccv2018,tian_lan14,zolf18}. 
Han et al. \cite{tengda17} introduced a stacked LSTM based method to learn the task grammar and predict future using both RGB and flow cues. The key component of their method is the estimation of task progress which considers separate networks for each level of granularity. This makes the approach not only inefficient but also very specific to each task since granularity level for different activities and environments is not the same. To predict the starting time and label of the future action, Mahmud et al. \cite{mahmud_iccv17} propose to use an LSTM to model long-term sequential relationships. More recently, Farha et al. \cite{whenwhat18} proposed a deep network to predict future activity. These methods rely on partial observations of the future and their predictions are limited to a fixed time horizons into the future. Another very interesting future prediction task required by autonomous driving, interactive agents or surveillance systems is forecasting the locations of objects or humans in the future \cite{Bhattacharyya17,chenyou_17}. Fan et al. \cite{chenyou_17} introduced a two-stream network to infer future representations to predict future locations. Their method is limited to 1 to 5 seconds into the future. Bhattacharyya et al. \cite{Bhattacharyya17} further addressed the multi-modality and the uncertainty of the future prediction by modeling both data and model uncertainties. 
    
\textbf{ Future Feature Prediction.} Prediction of future in semantic level is more easier and appealing for many applications such as autonomous driving \cite{NextSegmPredICCV17}. Vondrick et al. \cite{vondrick15} predicted visual representation of future frame. This approach is based on single frame an therefore is limited in terms of dynamics of the actions and also considers a short time horizon of 5 seconds into the future. 
    
In contrast to these works, we limit ourselves to only look at the current observations to infer the future activity without limiting the time horizon of the future prediction. Inspired by by Luc et al. \cite{paul_segmentation} we explicitly learn translation from current features to future features. Moreover, we address the ambiguous nature of the future by predicting multiple possible future representations with their uncertainties.

\textbf{Uncertainty Estimation in CNNs.} Modern CNNs are shown to be overconfident about their estimations~\cite{calibration} which makes them less trusted than non-blackbox traditional counterparts, despite their high performance. Recently, well-calibrated uncertainty estimation for CNNs has gained significant importance in order to tackle this shortcoming. 
One of the most popular uncertainty estimation methods for modern CNNs is MCDropout by Gal and Ghahramani~\cite{baysianDropout,visionUncertainties}. They show that using dropout over the weights for sampling, it is possible to get easy and efficient sampling for model uncertainty. Lakshminarayanan et al~\cite{deepEnsembles} propose using network ensembles over dropout ensembles for better uncertainty predictions. Another less resource expensive alternative is snapshot ensembling~\cite{snapshotEnsembles} over the networks trained with Stochastic Gradient Descent with Warm Restarts~\cite{sgdr}. All these methods still cannot avoid the sampling cost. Ilg et al~\cite{mhp} propose multi-hypotheses networks (MHN)~\cite{MultipleChoiceEnsembles,PhotographicImageSynthesis,rupprecht} for uncertainty estimation in order to overcome sampling. They show that the MHN is not only able to produce multiple samples in one forward pass but also provides with the state-of-the-art uncertainty estimates for optical flow. To this end, we modify the MHN for classification for our uncertainty estimation for future action classification.

%% file: method.tex
\section{Prediction of future activities}

Given an observation at a current time segment $t$, future activity prediction aims for estimating the activity class of the video segment at $t+ \Delta$, where $\Delta$ is the prediction horizon. Rather than learning directly this mapping, which we show to be clearly inferior, we use the features from an activity classification network for the current time segment and learn a mapping from these current features to features in the future. 

\begin{figure}[t]
\centering
\includegraphics[trim=0mm 0mm 0mm 0mm,clip,width=0.48\textwidth]{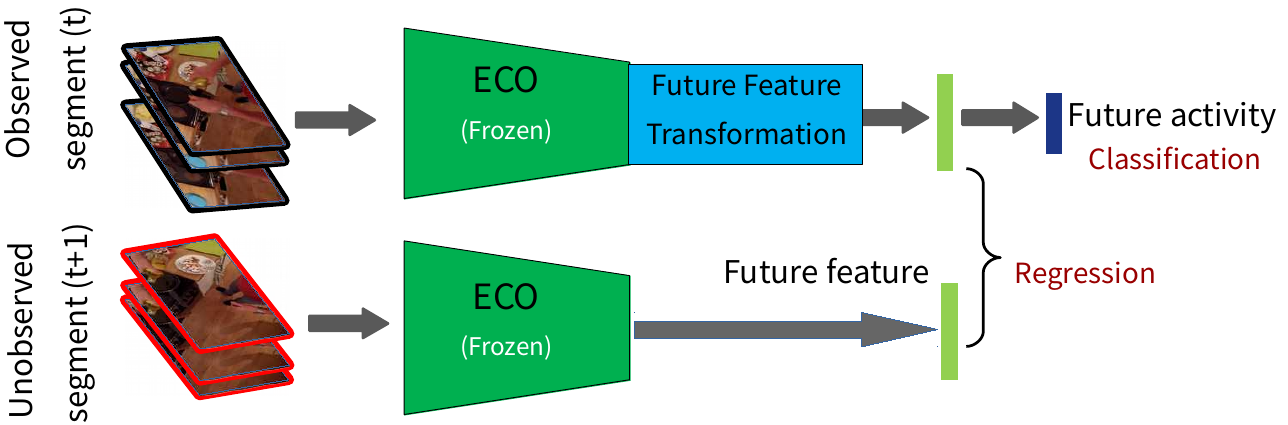}
\caption{Training scheme for PreFAct. The model learns to transform features from the observed input to future features based on a supervised classification loss and/or a self-supervised regression loss. 
}
\label{fig:train}
\end{figure}

\begin{figure*}[t]
\centering
\includegraphics[trim=0mm 0mm 0mm 0mm,clip,width=1\textwidth]{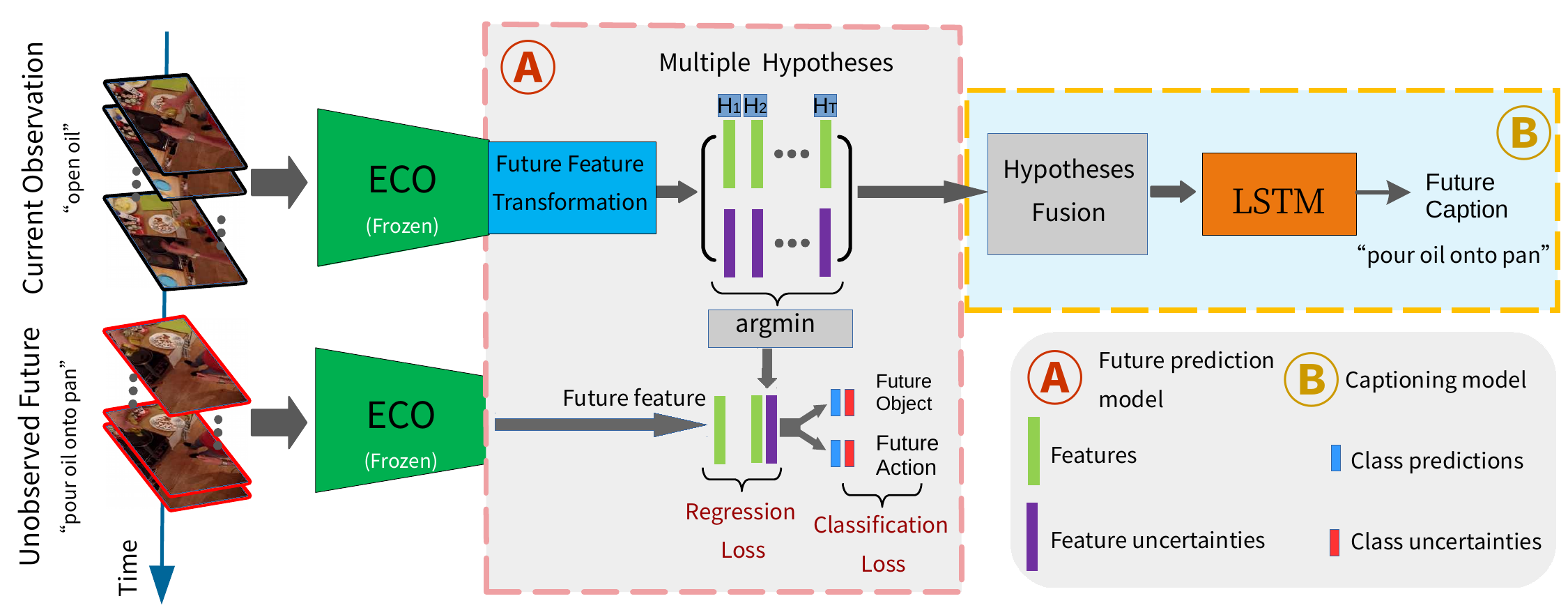}
\caption{ {\bf Overview of the future prediction model: PreFAct.} It consist of two main modules: Module "A" is the future representation learning stage. Given current observation, model learns to transform current features to the future representations and predict the labels of the future object and action. Module "B" is an LSTM based network which captions the future by fusing the multiple feature representations and their corresponding uncertainties. 
}
\label{fig:main_fig}
\end{figure*}

A coarse view of the network training is shown in Figure~\ref{fig:train}. Figure~\ref{fig:main_fig} shows a more detailed view of the overall model. As the base action classification network, we use ECO \cite{zolf18}. Between the convolutional encoder and the fully connected layers of ECO, we add the future prediction module $F_w$. We explore different designs for this module, shown in Figure~\ref{future_modules}. This also includes three different inception blocks. Moreover, we evaluate six different ways on where to include these modules into the ECO architecture, as illustrated in Figure~\ref{future_archs}. Experimental results with these different options are presented in Section~\ref{sec:modules}.

The weights of the ECO base network stay fixed (both the convolutional encoder and the fully connected activity classifier), while the future prediction module is trained. The training scheme is illustrated in Figure~\ref{fig:train}. Ground-truth features in the future are simply extracted by running ECO on the future time segment (lower part of Figure~\ref{fig:train}), i.e., training can work in a self-supervised manner as regression without annotation of class labels. The corresponding training objective is simply the mean squared error between predicted features $F_w(f^t_i)$ and extracted features $f^{t+\Delta}_i$ for video segment $i$:
\begin{equation} \label{eq2}
L_{reg} = \sum_{i}{||F_w(f^t_i)-f^{t+\Delta}_i||_2^2}. 
\end{equation}

Concurrently, the future prediction module can also take the class labels of a labeled video into account. In this case, the weights are optimized for the cross-entropy loss on the activity class labels $y^{t+\Delta}$ in the future time segment: 
\begin{equation} \label{eq3}
L_{class} = -\sum_{i}{y^{t+\Delta}_i\log(p^{t+\Delta}_i)}
\end{equation}
where $p^{t+\Delta}$ is the \textit{softmax} output for the future activity prediction based on $F_w(f^t_i)$.

\subsection{Class representation of objects and actions}
One way to represent the result of future prediction is by activity classification. Human actions are characterized by the \emph{objects} they interact with and the \emph{actions} they perform. In activity learning, often, one optimizes for the activity class directly, which leads to a combinatorial explosion of possible activities. Training directly on these can result in bad representations. For example, if in the training set the action 'put' is always combined with 'plate', the network will not learn the action 'put' but rather will recognize plates. Such representation will not generalize to somebody putting a cup.

Understanding the relationship between actions and objects leads to a more comprehensive interpretation. For instance, if the model already learned what a 'put' action means, it can more easily generalize to various scenarios such as 'put butter' or 'put spoon'. This enables us to extend the model to unseen objects-activity combinations by providing only a very small set of samples. Therefore, we propose to decouple the object and action classes. Treating them as separate sets but learning them jointly still exploits the relationship between them.
We will show the advantage of this decoupling in Section~\ref{sec:fewshot}.

\subsection{Video captioning}
A richer way of representing the results of the prediction module is via language. A video caption usually has more details than an activity class label. For instance, the caption `put celery back into fridge' conveys more details than the label `put celery'. 
We use an LSTM based architecture - semantic compositional networks \cite{SCN_CVPR2017} - for generating a caption describing the future feature representation. The semantic concepts are trained separately from scratch for each video dataset. These concepts are used to extend the weight matrices of caption generating decoder, as described in \cite{SCN_CVPR2017}


%% file: multiHypo.tex
\section{Multiple hypotheses and uncertainty}
\label{sec:multi_future_hyp}
If the next action is deterministic and depends only on the previous action, learning the mapping from the present to the future is almost trivial. It is a simple look-up table to be learned. However, the future action typically depends on subtle cues in the input and contains non-deterministic elements. Thus, multiple reasonable possibilities exist for the future activity. 
Therefore, we propose learning multiple hypotheses with their uncertainties, similar to multi-hypotheses networks (MHN)~\cite{mhp,MultipleChoiceEnsembles,PhotographicImageSynthesis,rupprecht}. In our setting, a multi-hypotheses network is used for predicting multiple feature vectors corresponding to the various possible outcomes together with their uncertainties. Each hypotheses yields the object and action class together with their class uncertainties; see Figure~\ref{fig:main_fig}.
We have separate uncertainties for objects and actions because each task has different uncertainty levels. For instance, if a person is washing and there is a spoon, a plate, and a knife in the sink, the uncertainty for the chosen object will be much higher than for the action. The feature uncertainties allows the captioning LSTM to reason about which features are most likely to rely on. 

To model the data uncertainty (aleatoric), the network yields the parameters of a parametric distribution, e.g., a Gaussian or Laplacian. This enables learning not only the mean prediction but also its variance, which can be interpreted as uncertainty. To cover the model uncertainty (epistemic), however, sampling from the network parameters is needed to compute the variation inherent within the model.
Multiple-hypotheses networks create multiple samples in one forward pass, which approximates sampling from the network in a very efficient way.

\begin{figure*}[t]
\centering
\includegraphics[trim=0mm 0mm 0mm 0mm,clip,width=1\textwidth]{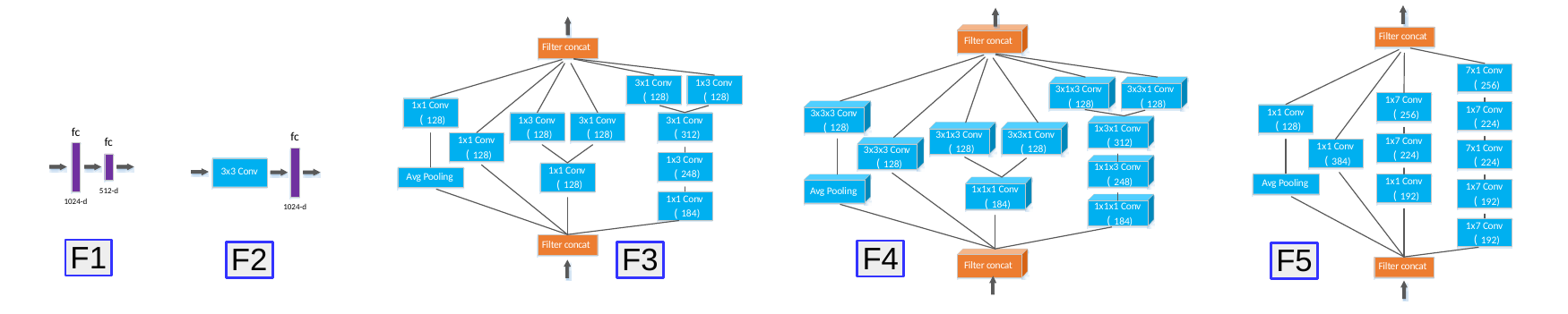}
\caption{ Future feature transformation modules. We considered fully-connected layers, convolutional layers, and different inception modules.}
\label{future_modules}
\end{figure*}

\begin{figure*}[t]
\centering
\includegraphics[trim=0mm 0mm 0mm 0mm,clip,width=1\textwidth]{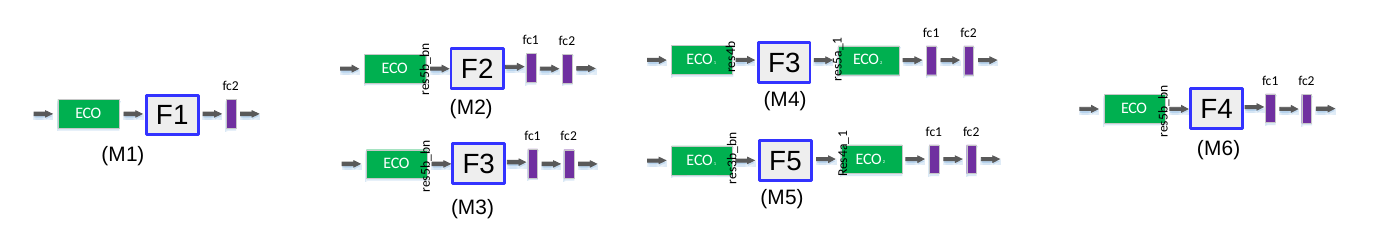}
\caption{Architectures for future representation learning. The transformation modules from Figure~\ref{future_modules} can be integrated into the ECO architecture at various places. We considered 6 variants shown in this figure.}
\label{future_archs}
\end{figure*}

\subsection{Feature uncertainties}

Following Ilg et al.~\cite{mhp}, we model our posterior by a Laplace distribution parameterized by median $a$ and scale $b$ for ground-truth feature $f^{gt}$ as:
\begin{equation} 
\mathcal{L}(f^{gt}|a,b) = \frac{1}{2b}e^{-\frac{|f^{gt}-a|}{b}} \mathrm{.}
\end{equation}

During training, we minimize its negative log-likelihood (NLL):
\begin{equation} 
{NLL}(f^{gt}|a,b) = \frac{\sqrt{(f^{gt}-a)^2}}{b}+{\log{b}} \mathrm{.}
\end{equation}
As commonly done in the literature, we predict $\log{b}$ instead of $b$ for more stable training. 

To also include the model uncertainty, we minimize the multi-hypotheses loss:
\begin{equation}
    L_{hyp} = \sum_{i} NLL(
    {f}_{\mathrm{best\_idx}}(i), 
    {f}^{\mathrm{gt}}(i)) \mathrm{.}
\label{eq:lhyp}
\end{equation}
For each training sample $i$, only the best feature among all hypotheses is penalized while the others stay untouched. The best feature is defined as the hypothesis closest to its ground-truth in terms of $L_2$ distance as follows:
\begin{equation}
    \mathrm{best\_idx} = \underset{k}{\mathrm{arg min}} \, \sum_{i}{||f_{k}(i) - {f(i)^ {gt}}||_2}
    \mathrm{.}
\label{eq:best_idx}
\end{equation}
This winner-takes-all principle, also used in \cite{mhp,MultipleChoiceEnsembles,PhotographicImageSynthesis,rupprecht}, fosters diversity among the hypotheses.

\subsection{Classification uncertainties}
For the classification loss, we model the data uncertainty as the learned noise scale~\cite{visionUncertainties}. In order to learn both the score maps and their noise scale, we minimize the negative expected log likelihood:
\begin{equation}
     L(s)=-\sum_{i}{\log\left[{\frac{1}{T}}\sum_{t}{\exp(\hat{s}_{i,t,c^{'}}-\log{\sum_{c}{\exp(\hat{s}_{i,t,c})}})}\right]}
\end{equation}

where $c^{'}$ is the observed class and $\hat{s}$ are predicted logits corrupted by Gaussian noise with the learned noise scale $\sigma$. Note that both $s$ and $\sigma$ are learned by the network. This formula can be interpreted as first corrupting the logits with noise $T$ times, where $T$ is the number of hypotheses, and normalizing them by softmax to get pseudo-probabilities $p_t(c)$, then averaging over these $T$ pseudo-probabilities to get the final pseudo-probabilities $p(c)$. Finally, the cross-entropy loss is applied to the pseudo-probabilities.

\textbf{From variances to uncertainties.} 
For both feature regression and object/action classification, we compute the final uncertainties as the entropy of the distributions:
\begin{equation}
 \mathcal{H}_{reg}(b) = \log{2eb},
\label{eq:ent1}
\end{equation}

\begin{equation}
    \mathcal{H}_{class}(p) = -\sum_{c}{p(c)\log{p(c)}}
\label{eq:ent2}
\end{equation}

%% file: results.tex
\section{Experiments}



\subsection{Datasets}

Since future activity prediction received little attention so far, there is no dedicated dataset for this task. We conducted our experiments on the Epic-Kitchens dataset~\cite{epickitchen_dataset} and the Breakfast dataset~\cite{breakfast_dataset}. Both show sequential activities on preparing meals with sufficient diversity. These two are the most suitable datasets for our task since they include temporally meaningful actions which follow each other in a procedural way i.e. "Peel Potato" is followed by "Cut Potato". 



\textbf{Epic-Kitchens}\cite{epickitchen_dataset}
 dataset includes videos of people cooking in a kitchen from a first person view. Each video is divided in multiple video segments. In total there are $272$ video sequences with $28,561$ activity segments for training/validation and $160$ video sequences with $11,003$ action segments for testing. These segments are annotated using in total $125$ verb and $331$ noun classes.
 From the video sequences in training/validation dataset we randomly choose 85\% of the videos for training and 15\% of them for validation.

\textbf{Breakfast}\cite{breakfast_dataset}
 dataset includes meal preparation videos of $10$ common breakfast items in third person view. On average each item has $200$ preparation videos where some videos are from the same scene with multiple camera angles. All videos are divided in multiple video segments which are annotated with one of $48$ predefined activity labels. We convert activity classes into object and action classes, e.g.  activity: "take cup" becomes action: "take" and object: "cup".
 All videos in this dataset are provided by 52 participants. We use data from 39 of these participants for training and data from remaining 13 participants for testing.

\subsection{Evaluation metrics}

For classification, we use accuracy as quantitative measure, i.e. the rate of correctly predicted classes over the whole predictions.

The captioning models are evaluated using the standard metrics BLEU (B-1) \cite{papineni2002bleu}, ROUGE\_L\cite{lin2004rouge}, METEOR \cite{michael2014meteor}, and CIDEr \cite{vedantam2015cider}.
BLEU (B-1) calculates the geometric mean of n-gram precision scores weighted by a brevity penalty. 
ROUGE\_L measures the longest common sub-sequence between generated caption and the ground-truth. 
METEOR is defined as the harmonic mean of precision and recall of matched uni-grams between generated caption and its ground-truth. 
CIDEr measures the consensus between generated caption and the ground-truth.

For evaluating the quality of the uncertainty predictions we use reliability diagrams~\cite{calibration}. A reliability diagram plots the expected quality as a function of uncertainty. If the model is well-calibrated this plot should draw a diagonal decrease.



\subsection{Implementation and training details}


We base our feature extraction module on ECO~\cite{zolf18}. Following the original paper, we take the ECO which was pretrained on Kinetics~\cite{kineticsDataset} and then further trained on Breakfast or Epic-Kitchens depending on the dataset used in the experiments. When we retrain the ECO for the baseline comparisons we follow the design choices of the original paper as is if not mentioned otherwise. We provide all details in supplementary material. Data augmentation is also applied as in the original work. Keeping the ECO feature extractor fixed, we train our future representation module which is initialized randomly. We use mini-batch SGD optimizer with Nesterov momentum of $0.9$, weight decay of $0.0005$, and mini-batches of $64$. We utilize dropout after each fully connected layer. For the multi-hypotheses experiments we fix the number of hypotheses (T) to $8$.

     


We extract frames from the video segments following the sampling strategy explained in the original paper. In this sampling, each segment is splitted into 16 subsections of equal size and from each subsection a single frame is randomly sampled.
This sampling provides robustness to variations and enables the network to fully exploit all frames and enable us to predict arbitrary horizons into the future.  


\subsection{Comparison of feature translation modules}
\label{sec:modules}
Table \ref{tab:featuremodulecompare} compares different design choices for the feature translation module, as depicted in Figures~\ref{future_modules},\ref{future_archs}. The architectures M3 and M6 provide the best performance. M3 corresponds to locating a $8\times8$ grid inception block with 2D convolutions before the last two fully connected layers. M6 is the same with 3D convolutions. For the rest of the experiments, we used M3 as a feature transformation module.
In these experiments, we report the accuracy on the (composed) activity recognition for the Breakfast dataset and on the (single) action recognition for the Epic-Kitchens dataset. This differs from the results we provide in the following sections where we evaluate our decomposed action/object classes separately.


\begin{table}
\begin{center}
\caption{Comparison of design choices for the feature transformation modules. M3 and M6, as shown in Figures~\ref{future_modules},\ref{future_archs}, are best.}
\label{tab:featuremodulecompare}
\begin{tabular}{lcccccc}
\hline\noalign{\smallskip}
Dataset &  M1 &  M2 & M3 &  M4 &  M5 & M6  \\
\noalign{\smallskip}
\hline
\noalign{\smallskip}
Breakfast             & 28.1        & 27.7     & \textbf{28.9}  & 24.4 & 19.4  & 28.4  \\
Epic-Kitchens    & 29.5 & 31.2  & 31.4 & 28  & 28.8 & \textbf{32.4}    \\
\hline\\[-10mm]
\end{tabular}
\end{center}
\end{table}

\begin{table}[t]
\begin{center}
\caption{Next activity classification accuracy on the Breakfast (Brk) and Epic-Kitchens (Epic) datasets. "A": Action, "O": Object. PreFAct improves over all baselines, also the ECO baseline, which corresponds to learning the mapping from observation to future class label directly. }
\label{tab:futureactivity_acc}
\begin{tabular}{lcccc}
\hline\noalign{\smallskip}
Methods &  Brk (A) & Brk (O) & Epic (A) & Epic (O)  \\
\noalign{\smallskip}
\hline
\noalign{\smallskip}
\color{blue} Current Activity \\ \color{blue}(Upper bound)   & \color{blue} 68.0 & \color{blue} 51.4 & \color{blue} 56.3 & \color{blue} 30.8     \\ \hline

Largest class   & 22.5 & 12.4 & 20.67 & 4.44     \\
Copy current label   & 8.8 &  9.6 & 21.32 & 13.92     \\
Assoc. rule mining  & 16.1 & 8.7 & 27.9 & 14.2      \\
Epic-Kitchens\cite{epickitchen_dataset}   & 31.7  & 25.2 & 30.6 & 15.8    \\ 
ECO\cite{zolf18}   & 38.9  & 35.4 & 30.3 & 13.2    \\ \hline
PreFAct-C   & 41.4 & 36.3 & 32.0  & 15.6     \\
PreFAct-R   & 38.2 & 33.3 & 29.7 & 14.9   \\
PreFAct-R+C   & 41.8 & 37.2 & 32.2 & 15.8    \\ \hline
PreFAct-MH_{Best}   & 42.1 & 36.4 & 32.0 & 15.7    \\ 
PreFAct-MH_{Oracle}   & 47.4 & 40.9 & 37.7 & 18.4  \\  
\hline\\[-10mm]
 \end{tabular}
\end{center}
\end{table}

\subsection{Results on future activity prediction}
\label{sec:expclass}

Due to the lack of previous work on this problem, we compare to some simple and some more in-depth baselines. A comparison of these baselines is shown in Table~\ref{tab:futureactivity_acc}. The table provides as upper bound the classification accuracy with ECO where the time frame of interest is observed, i.e., the problem is a standard classification problem without a future prediction component. 

The "largest class" baseline just assigns the label of the largest class in the training data to the label of the future activity. This is the accuracy achieved by simply exploiting the data imbalance of the datasets. 

The "copy current label" baseline performs activity classification on the current observation and considers the predicted current label as the future activity label. This approach only works in cases, where the action or the object do not change over time. 

The "association rule mining" picks the most likely future activity as the label of the future activity. See supplementary material for details about each baseline.

As can be seen from the Table, the results we get with the future activity prediction network (PreFAct) are much better than these simple baselines. PreFAct-C denotes the network that was only trained using the class labels supervisedly, whereas PreFAct-R was trained trained only in a self-supervised manner without using class labels. PreFAct-R+C used both losses jointly for training. As expected, supervised training works better than only self-supervised training, and using both losses works marginally better than only supervised training. Note that the self-supervised learning can leverage on additional unlabeled video data. We explore this more in detail in Section~\ref{sec:selfsupervised}. 

The two most interesting baselines are the two state-of-the-art methods on video understanding - ECO~\cite{zolf18} and Epic-Kitchens~\cite{epickitchen_dataset} - which we trained to predict future activities rather than the present activity. For a fair comparison, we modified the methods such that they provide both object and action classes rather than a single activity class. PreFAct clearly improves over this ECO baseline, which shows that the future prediction module is advantageous over directly learning the mapping from the observation to the future activity. 

PreFAct-MH shows results with our multi-hypotheses network.
Generating multiple hypotheses has potential to lead to significant performance improvement, as demonstrated by the Oracle selection, where the best hypothesis is selected based on the true label. Whereas,
automated selection of the best hypothesis via their uncertainty estimates does not lead to a significant difference over the version with single hypothesis. This is consistent with the findings in other works, which showed good uncertainty estimates, but could not benefit from these hypotheses to select the best solution within a fusion approach.



\subsection{Learning unseen combinations}
\label{sec:fewshot}

The decomposition of activities into the action and the involved object allows us to generalize to new combinations not seen during training. Table~\ref{tab:zeroshot} shows results on an object-action pair when all pairs of the specified object with the 5 actions in the top row were completely removed from the training data. In brackets are the numbers when these activities were part of the training set. In most cases, the approach is able to compensate for the missing object-action pairs by using the information from a related object or another action not among the 5 actions.

\begin{table}
\begin{center}
\caption{Action prediction of current observation for unseen object-action combinations on the Epic-Kitchens dataset. For each object-action pair, we report the accuracy when excluding all 5 actions for these objects during training. The numbers in parenthesis indicate the accuracy when training on the entire dataset. Our decomposition of action/object classes helps compensating for the missing information.}

\label{tab:zeroshot}

\begin{tabular}{lccccccc}
\hline\noalign{\smallskip}
 &  Take &  Put & Open &  Close &  Wash   \\
\noalign{\smallskip}
\hline
\noalign{\smallskip}
Cupboard   & -       & -      & 91 (94)   & 83 (91)  & -       \\
Drawer     & 3 (6)       & 3 (8)      & 94 (93)   & 70 (74)  & -      \\
Fridge    & -       & -      & 95 (97)   & 94 (97)  & -    \\
Plate    & 68 (70)       & 69 (69)      & -   & -  & 90 (90)       \\
Knife    & 69 (67)       & 69 (69)      & -   & -  & 97 (92)        \\
Spoon    & 65 (67)       & 70 (59)      & -   & -  & 100 (93)        \\
Lid    & 69 (80)       & 55 (60)      & 0 (50)   & 0 (50)  & 75 (80)      \\
Pot    & 57 (55)       & 71 (70)      & -   & -  & 80 (92)       \\

\hline
\end{tabular}
\end{center}
\end{table}

\subsection{Self-supervised learning}
\label{sec:selfsupervised}

While the self-supervised regression loss yields inferior results compared to supervised training on class labels, self-supervised learning has the advantage that it can be run effortlessly on unlabeled video data. 


Table~\ref{tab:unsupervised} shows how the self-supervised learning improves when adding extra unlabeled data S1 and S2 provided by the Epic-Kitchens dataset. S1 contains 8048 samples of seen kitchens, and S2 contains 2930 samples of unseen kitchens. The improvement is small but increases as more data is added.    

\begin{table}
\begin{center}
\caption{Performance of the self-supervised learning (PreFAct-R) on the Epic-Kitchens dataset (left/right: object/action), as additional unlabeled data S1 or S2, or both is added for training. The more the unlabeled data the better.}
\label{tab:unsupervised}
\begin{tabular}{lccccccc}
\hline\noalign{\smallskip}
Method &  None &  S1 & S2  & S1+S2  \\
\noalign{\smallskip}
\hline
\noalign{\smallskip}
PreFAct-R             & 14.9/29.7         & 15.1/30.1 & 15.0/30.3 & 15.3/30.5 \\
\hline\\[-10mm]
\end{tabular}
\end{center}
\end{table}


\subsection{Future captioning}
\label{sec:future_caption}

\begin{figure}[t]
\centering
\includegraphics[trim=0mm 0mm 0mm 0mm,clip,width=0.49\textwidth]{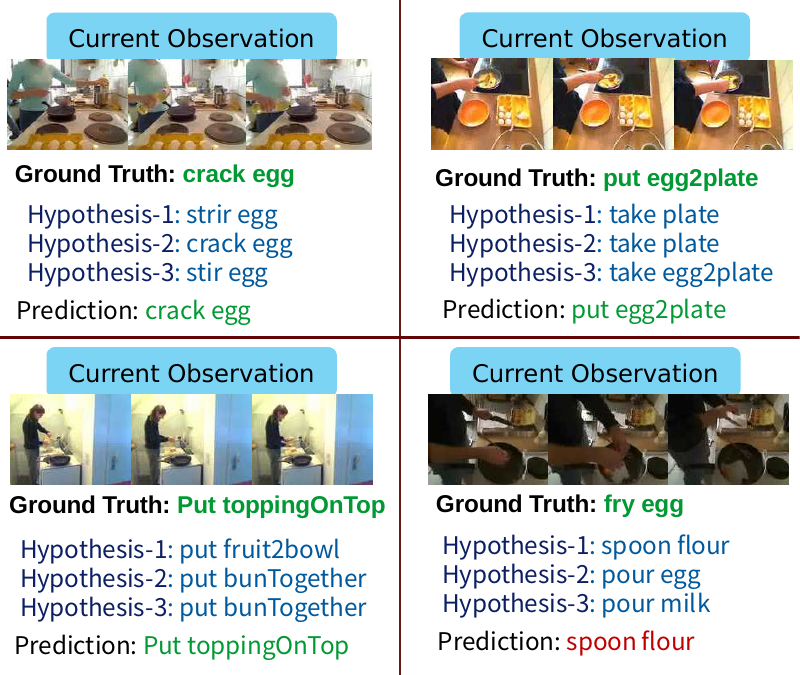}
\caption{ { Qualitative results on Breakfast dataset obtained from our captioning module with $All_{mult}$ feature fusion - only TOP3 Hypotheses are visualized.}}
\label{fig:qualitative_results1}

\end{figure}

We use semantic compositional networks \cite{SCN_CVPR2017} for captioning current and future video features. For each dataset, we obtain a separate semantic concept detector by training a multi-label classifier for the set of selected concepts from each dataset. For most experiments, we use the full vocabulary as the set of concepts.  

Our feature representations provide multiple features and classes with their uncertainties. We explored various options on how to fuse and feed this information into the captioning module (Fig.~\ref{fig:main_fig}(B)). They are tagged as ${\rm SELECTION}_{how\_to\_fuse}$ in Table~\ref{tab:feature_fusion}. We use the class certainties to \textit{select} features: feature yielding the highest class certainty (BEST), the highest three certainty (TOP3), and all features (ALL). For \textit{fusion} of the selected features, we considered: concatenation of them with their certainties (concat), and multiplying them with their certainties and concatenating (mult). We obtain the certainties by first normalizing the uncertainties to (0,1) range and then subtracting them from 1.

Table \ref{tab:feature_fusion} compares these different options. Using all feature hypotheses magnified with their certainties ($All_{mult}$) yield the best results with large margin in comparison to other alternatives. This suggests that capturing the future with its multi-modality and variation is the key to represent future semantics. 

Table~\ref{tab:captioning1} and \ref{tab:captioning2} shows that the multi-hypotheses design is clearly superior to its single prediction counterparts on both the Breakfast and the Epic-Kitchens dataset. While multiple hypotheses could not be exploited at the classification level in Section~\ref{sec:expclass}, they help a lot on the captioning task.

Figure~\ref{fig:qualitative_results1} shows some qualitative results of future captioning on the Breakfast dataset. For each sample, future action/object classes of top-3 hypotheses are presented. In the top-left case, hypotheses are certain about the future object "egg", but for the action there is high uncertainty. In contrast, in the bottom-left case, uncertainty on the future object is higher than for the future action "put". 

\subsection{Uncertainty evaluation}
\begin{figure}[t]
\centering
\includegraphics[trim=0mm 0mm 0mm 0mm,clip,width=0.49\textwidth]{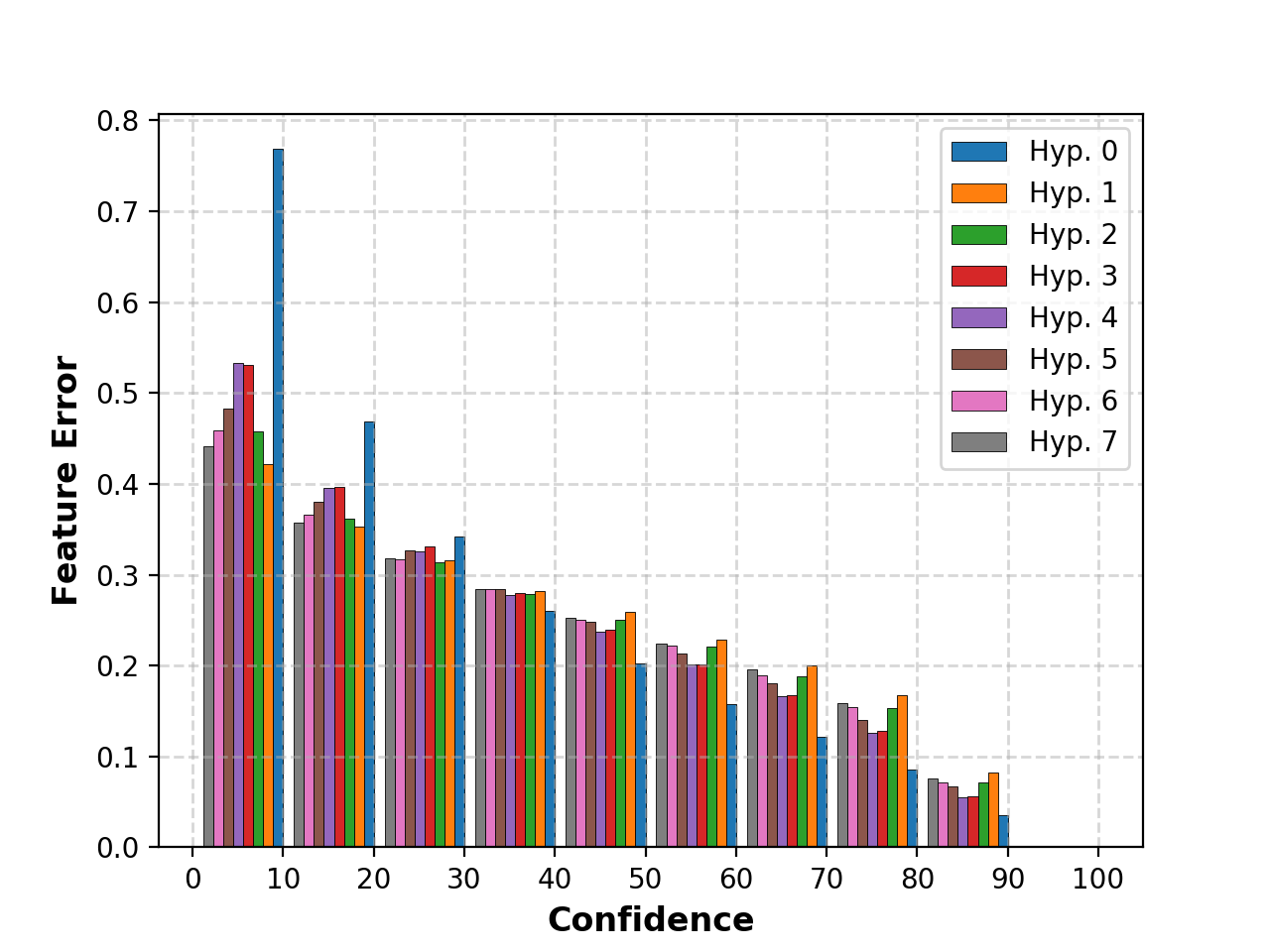}
\caption{ Reliability diagram for our feature uncertainties on Epic-Kitchens dataset. Diagonal decrease suggests that our uncertainties are well calibrated and potentially useful. }
\label{fig:bin}
\end{figure}

In Figure \ref{fig:bin}, we provide the reliability diagram for the uncertainties of feature hypotheses for Epic-Kitchens dataset. The diagonal decrease suggests that our uncertainties are well calibrated with the errors of the features. In the supplementary material we provide more details about our method and more in-depth evaluations as well as more qualitative results including failure cases.

\begin{table}

	\begin{center}
		\begin{tabular}{c|c|cccc|}
			\cline{2-6}
			\multirow{2}{*}{} & {Hypotheses}&  \multicolumn{4}{c|}{Captioning performance }\\
			\cline{3-6}
			& fusion & {CIDEr}& {B-1}&{Rouge\_L}& {METEOR}\\
			\cline{2-6}
        \multirow{8}{*}{\rotatebox{90}{}}  
                                               & BEST_{concat}       & 0.270  & 0.173 & 0.171 & 0.084 \\
                                               & BEST_{mult}        & 0.372  & 0.233 & 0.218 & 0.107  \\\cline{2-6}
                                               & TOP3_{concat}      & 0.294  & 0.186 & 0.185 & 0.091 \\ 
                                               & TOP3_{mult}        & 0.350  & 0.221 & 0.210 & 0.103 \\ \cline{2-6}

                                               & ALL_{concat}      & 0.317  & 0.201 & 0.195 & 0.096 \\ 
                                               & ALL_{mult}        & \bf{0.466}  & \bf{0.294} & \bf{0.269} & \bf{0.132} \\ \cline{2-6}

            \cline{2-6}
		\end{tabular}
	\end{center}
	\caption{Captionining performance based on different hypotheses selection and processing strategies. For "Top3" and "Best" we pick the feature with the lowest $3$ and $1$ class uncertainty respectively. Lower entropy means that we are more certain about the prediction. "All" means all features are fed to the captioning. "concat": concatenation of feature and certainty vectors, "mult": multiplying features with their certainties and concatenating.}
	\label{tab:feature_fusion}

\end{table}

\begin{table}[t]
\small
  \begin{center}
    \begin{tabular}{lcccccccccccccccc}
      \toprule

      {} & \multicolumn{4}{c}{Metrics} \\

      \cmidrule{2-5} 
      {Methods} & {CIDEr} & {B-1}&{Rouge\_L}& {METEOR}&  \\
      \midrule

     \color{blue} Current Activity  	    & \color{blue}  0.791        & \color{blue}  0.503       & \color{blue}  0.454           &   \color{blue}  0.223     \\
     ECO\cite{zolf18}        & 0.319   & 0.206       & 0.201           &   0.099       \\ \hline
     PreFAct_{RC}           & 0.357    & 0.224     & 0.213          &   0.105      \\
     PreFAct_{MH}         & \bf{0.466}  & \bf{0.294} & \bf{0.269} & \bf{0.132}       \\

      \bottomrule
    \end{tabular}
  \end{center} 
  \caption{
  Future captioning results on Breakfast dataset. PreFAct$_{RC}$: Our model with regression and classification losses, PreFAct$_{MH}$: Our Multi-Hypotheses model.}
  \label{tab:captioning1}
  
\end{table}

\begin{table}[t]
\small
  \begin{center}
    \begin{tabular}{lcccccccccccccccc}
      \toprule

      {} & \multicolumn{4}{c}{Metrics} \\

      \cmidrule{2-5} 
      {Methods} & {CIDEr}& {B-1}&{Rouge\_L}& {METEOR}&   \\
      \midrule

     \color{blue}  Current Activity	   & \color{blue}  0.465  & \color{blue}  0.360       & \color{blue} 0.362           &   \color{blue}  0.205       \\
     ECO\cite{zolf18}       & 0.099  & 0.068       & 0.086           &   0.048         \\ \hline
     PreFAct_{RC}        & 0.145   & 0.105      & 0.113           &   0.065         \\
     PreFAct_{MH}        & \bf{0.165}  & \bf{0.122}      & \bf{0.138}          &   \bf{0.077}         \\
      \bottomrule
    \end{tabular}
  \end{center} 
  \caption{
  Future captioning results on the Epic-Kitchens dataset. PreFAct$_{RC}$: Our model with regression and classification losses, PreFAct$_{MH}$: Our Multi-Hypotheses model.}
  \label{tab:captioning2}
  
\end{table}

%% file: conclusion.tex
\section{Summary}
We presented the problem of predicting future activities based on past observations. To this end, we leveraged a feature embedding used for action classification and extended it by learning a dynamics module that transfers these features into the future. The network predicts multiple hypotheses to model the uncertainty of the future state. While this had little effect on future activity classes, it helped substantially for future video captioning. Due to the decomposed representation into object and action, the approach generalizes well to unseen activities. The approach also allows for fully self-supervised training. Although the performance is still inferior to the supervised setting, it has a lot of potential when applied to large-scale unlabeled videos. We believe there is promise in investigating further in this direction.

%% file: sup_mat.tex
\section{Baseline: Association rule mining}
\label{sec:a_rule}





 "Association rule mining" \cite{Agrawal_arm} discovers relations between activities in the dataset. For instance, in the Epic-Kitchens dataset actions "take" and "put" are occurred together frequently. Therefore, by identifying these relations we can predict the future class labels based on the current observation class label. 
  In the previous example, the rule would be:\\
  \emph{\textbf{If action "take" is observed, then "put" will be the next action}}.

 Using this method, we find the most probable patterns between activities. We first obtain the activity label ($y^{t}$) of the current observation and then, using association rule mining, the label of the future activity ($y^{t+ \Delta}$) will be the most co-occurred consequent activity ($y^{t} \rightarrow y^{t+ \Delta}$).

 \begin{table*}
 \small
   \begin{center}
   \caption{
   Frequency of action sequences in the Epic-Kitchens dataset. Support measures frequency of itemset, Confidence shows the probability of seeing the consequent action given previous action, and Lift measures how much dependent are two actions. }
   \label{tab:associate_rule}
   
     \begin{tabular}{lccc||lccc}
       \toprule

       \multicolumn{4}{c}{Actions} & \multicolumn{4}{c}{Objects} \\

       \cmidrule{1-8}  
        Itemsets & Support & Confidence & Lift & Itemsets & Support & Confidence & Lift \\
       \noalign{\smallskip}

       \midrule
       \noalign{\smallskip}

       $(take \rightarrow mix)$     & 0.010       & 0.027    & 0.502  &$(hand \rightarrow tap)$    &0.012  &0.299 &3.620 \\
       $(put \rightarrow wash)$     & 0.045       & 0.237    & 0.616  &$(plate \rightarrow cupboard)$  &0.0053  &0.089 &1.280 \\
       $(take \rightarrow put)$     & 0.122       & 0.325    & 0.844  &$(spoon \rightarrow drawer)$    &0.0045  &0.0811 &1.669 \\
       $(open \rightarrow put)$     & 0.0456      & 0.242    & 0.628  &$(sponge \rightarrow tap)$    &0.0049  &0.1834 &2.217 \\
       $(open \rightarrow take)$    & 0.0563      & 0.298    & 0.792  &$(air \rightarrow bag)$    & 0.0001  &1.00 &46.45 \\
       $(apply \rightarrow take)$   & 0.002       & 0.410    & 1.089  &$(dish:soap \rightarrow plate)$    &0.0003  &0.833 &13.75 \\
       $(hold \rightarrow mix)$     & 0.010       & 0.322    & 5.917  &$(mint \rightarrow leaf:mint)$    &0.0001  &0.2307 &433.84 \\
       $(spray \rightarrow wash)$   & 0.001       & 0.452    & 2.378  &$(shirt \rightarrow sock)$    &0.0002  &0.240 &376.00 \\
       $(cut \rightarrow peel)$     & 0.002       & 0.047    & 3.529  &$(mat:sushi \rightarrow omelette)$    &0.0001  &0.2857 & 537.14 \\

       \bottomrule
     \end{tabular}
   \end{center} 
 \end{table*}

 Table~\ref{tab:associate_rule} shows frequently occurring action sequences in the Epic-Kitchens dataset. In this table, we have provided three different components \emph{"Support"}, \emph{"Confidence"} and \emph{"Lift"}. \emph{Support} refers to the popularity of action set, \emph{Confidence} refers to the likelihood that an action "B" happens if action "A" is happened already, and \emph{Lift} measures dependency of actions.

 \begin{table*}
\resizebox{\textwidth}{!}{
\begin{tabular}{|c|c|c||c|c|}
\hline

F1 & F2 & F3 & F4 &F5   \\ 
\hline
FC [1024-d] & $ \begin{bmatrix} \mathrm{Conv} & 1 \times 3 \times 3  & 512 \end{bmatrix} $

& $\begin{bmatrix} IN: \mathrm{Conv} & 1 \times 1 \times 1 & 184:O1 \\  IN:\mathrm{Conv} & 1 \times 1 \times 1 & 128:O2 \\ IN:\mathrm{Conv} & 1 \times 1 \times 1 & 128:O3 \\ IN:\mathrm{Avg \ Pool} & 3 \times 3 \times 3 &:O4 \end{bmatrix}  $      

& $\begin{bmatrix} IN:\mathrm{Conv} & 1 \times 1 \times 1 & 184:O1 \\  IN:\mathrm{Conv} & 1 \times 1 \times 1 & 184:O2 \\ IN:\mathrm{Conv} & 3 \times 3 \times 3 & 128:O3 \\ IN:\mathrm{Avg \ Pool} & 3 \times 3 \times 3 &:O4  \end{bmatrix}  $    

& $\begin{bmatrix} IN:\mathrm{Conv} & 1 \times 1 \times 7 & 192:O1 \\  IN:\mathrm{Conv} & 1 \times 1 \times 1 & 192:O2 \\ IN:\mathrm{Conv} & 1 \times 1 \times 1 & 384:O3 \\ IN:\mathrm{Avg \ Pool} & 3 \times 3 \times 3 &:O4  \end{bmatrix}  $  \\

\hline

FC [512-d] & FC [512-d] & $\begin{bmatrix} O1:\mathrm{Conv} & 1 \times 1 \times 3 & 248:O5\\  O2:\mathrm{Conv} & 1 \times 3 \times 1 & 128:O6 \\ O2:\mathrm{Conv} & 1 \times 1 \times 3 & 128:O7 \\ O4:\mathrm{Conv} & 1 \times 1 \times 1 & 128 :O8  \end{bmatrix}  $  

& $\begin{bmatrix} O1:\mathrm{Conv} & 1 \times 1 \times 3 & 248:O5 \\  O2:\mathrm{Conv} & 3 \times 1 \times 3 & 128:O6  \\ O2:\mathrm{Conv} & 3 \times 3 \times 3  & 128:O7 \\ O4:\mathrm{Conv} & 3 \times 3 \times 3 & 128:O8 \end{bmatrix}  $ 

& $\begin{bmatrix} O1:\mathrm{Conv} & 1 \times 1 \times 7 & 192:O5 \\  O1:\mathrm{Conv} & 1 \times 1 \times 7 & 224:O6  \\ O1:\mathrm{Conv} & 1 \times 1 \times 1  & 128:O7  \end{bmatrix}  $  \\

\hline

~   & ~  &  $\begin{bmatrix} O5:\mathrm{Conv} & 1 \times 3 \times 1 & 312:O9  \end{bmatrix}$ & $\begin{bmatrix} O5:\mathrm{Conv} & 1 \times 3 \times 1 & 312:O8  \end{bmatrix}$ 

& $\begin{bmatrix} O1:\mathrm{Conv} & 1 \times 7 \times 1 & 224:O8 \\  O1:\mathrm{Conv} & 1 \times 1 \times 7 & 224:O9  \end{bmatrix}  $    \\

\hline

~ & ~ & $\begin{bmatrix} O9:\mathrm{Conv} & 1 \times 1 \times 3 & 128:O10 \\ O9:\mathrm{Conv} & 1 \times 3 \times 1 & 128:O11 \end{bmatrix}$  

& $\begin{bmatrix} O8:\mathrm{Conv} & 3 \times 1 \times 3 & 128:O9 \\ O8:\mathrm{Conv} & 3 \times 3 \times 1 & 128:O10  \end{bmatrix}$ &  

$\begin{bmatrix} O1:\mathrm{Conv} & 1 \times 1 \times 7 & 224:O10  \end{bmatrix}  $  \\

\hline

~&~ &$Concat(O3,O6,O7,O8,O10,O11):OUT$ & $Concat(O3,O6,O7,O8,O9,O10):OUT$

& $\begin{bmatrix} O1:\mathrm{Conv} & 1 \times 7 \times 1 & 256:O11  \end{bmatrix}  $   \\

\hline

~ &  ~ & ~  &$-$ & $Concat(O3,O7,O9,O10):OUT$   \\

\hline
\end{tabular}
}
\caption{{\bf Architecture details for feature translation modules}. The input to each module is the output of a specific layer of ECO network depicted in Fig. 5 of the main paper. }
\label{tab: FM_arch}
\end{table*}

 As shown in Table~\ref{tab:associate_rule}, the frequent action set is $(take \rightarrow put)$ which happens $12.2\%$ in the dataset and the frequent object set is $(hand \rightarrow tap)$ with $1.2\%$ occurrence. 
 We utilized \emph{Confidence} to find the most probable action "B" after observing action "A". For instance, if the current observed action is "Open" then action "put" will happen with probability of $24.2\%$.




\section{Implementation and training details}

During training, we use the SGD optimizer with Nesterov momentum of $0.9$ and weight decay of $0.0005$. Training is performed up to $80$ epochs for Epic-Kitchens dataset with randomized minibatches consisiting of $64$ samples, where each sample contains $16$ frames of a current video segment. 

For the Epic-Kitchen dataset, initial learning rate is $0.007$ and decreases by a factor of $10$ when validation
error saturates for $10$ epochs. 

Training is performed up to $60$ epochs for Breakfast dataset. We use dropout of $0.3$ for the last fully connected layer. For the Breakfast dataset, initial learning rate is $0.001$ and decreases by a factor of $10$ when validation
error saturates for $8$ epochs. 
In addition, we apply the data augmentation techniques similar to \cite{zolf18}: we resize the input frames to $240 \times 320$ and employ fixed-corner cropping and scale jittering with horizontal flipping.
Afterwards, we run per-pixel mean subtraction and resize the cropped regions to $224 \times 224$. 

During the inference time, we sample $16$ samples from the video, apply only center cropping and then feed them directly to the network to get final future predictions. For the captioning,  we utilize same approach but extracting the features from the regression layer. We use extracted features to train the LSTM to provide caption for each video segment.

\section{ Architecture details of feature translation modules}

\label{sec:modules}



For the future translation modules, we design several different architectures consist of fully connected layers and convolutional layers, see Table~\ref{tab: FM_arch}. For the $F3$, $F4$ and $F5$, we make use of inception modules introduced in \cite{SzegedyIV16}. For simplicity, we present each layer of modules $F3$, $F4$ and $F5$ in the following format (Table~\ref{tab: FM_arch}):\\
$[input: operation~~ d*h*w~~ \#filters:output]$
\\

\textbf{IN:} Input from a specific layer of the ECO network. \\

\textbf{Out:} Output to the rest of the ECO network.

\begin{figure}
\subfloat[Breakfast-Object\label{fig_sparsification:a}]{\includegraphics[width = 1.8in]{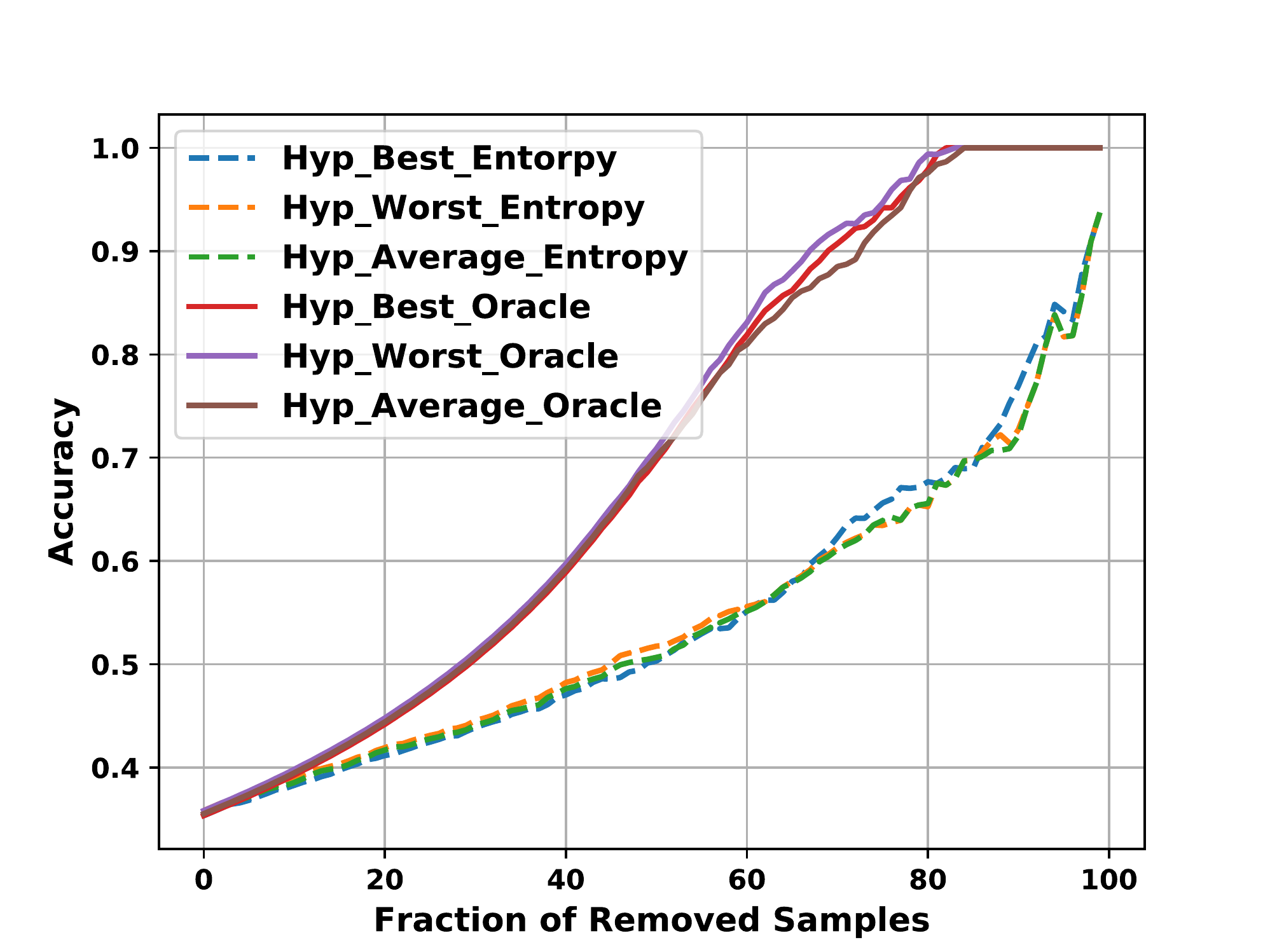}} 
\subfloat[Breakfast-Action\label{fig_sparsification:b}]{\includegraphics[width = 1.8in]{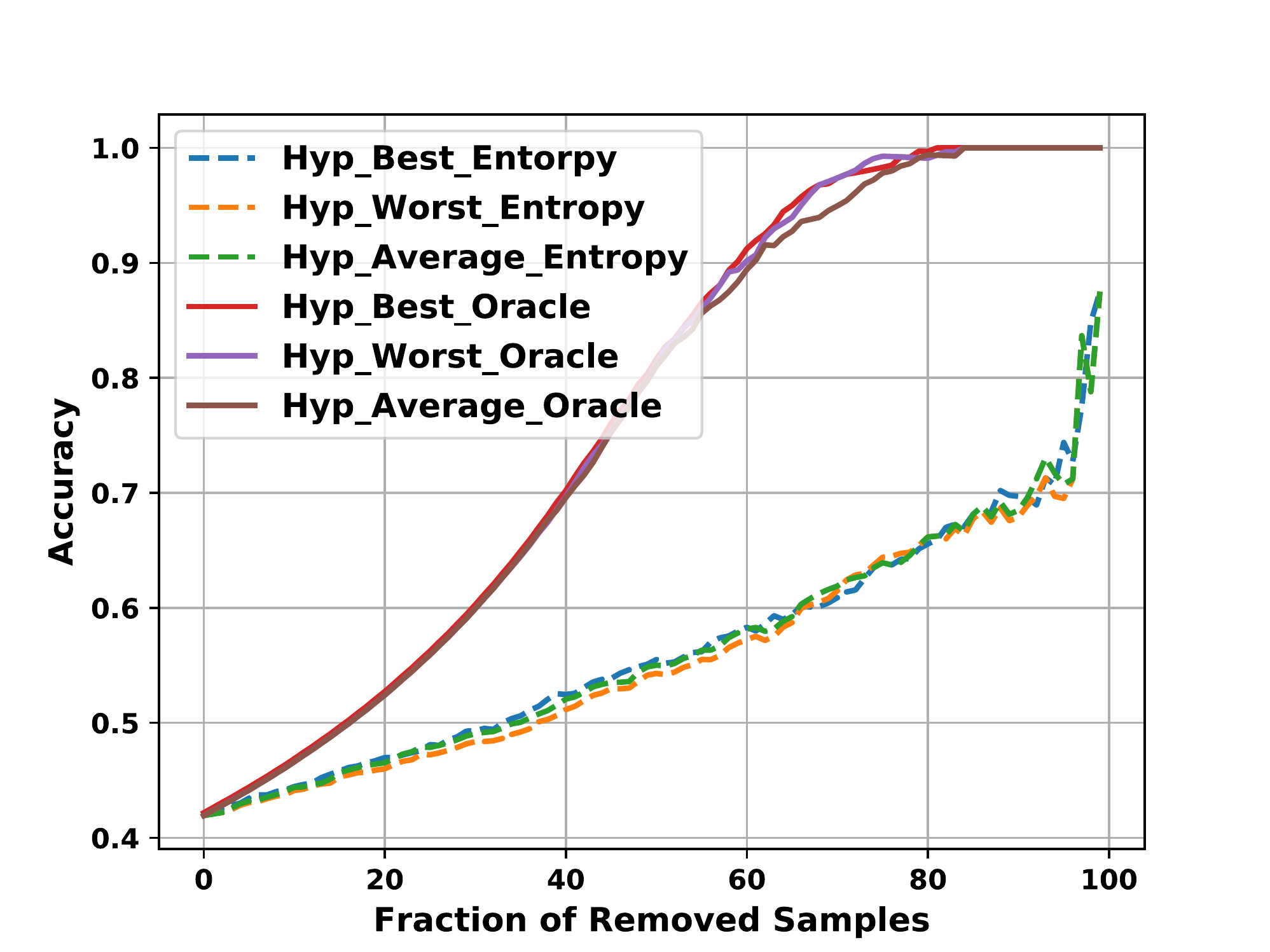}} \\ 
\subfloat[EpicKitchen-Object\label{fig_sparsification:c}]{\includegraphics[width = 1.8in]{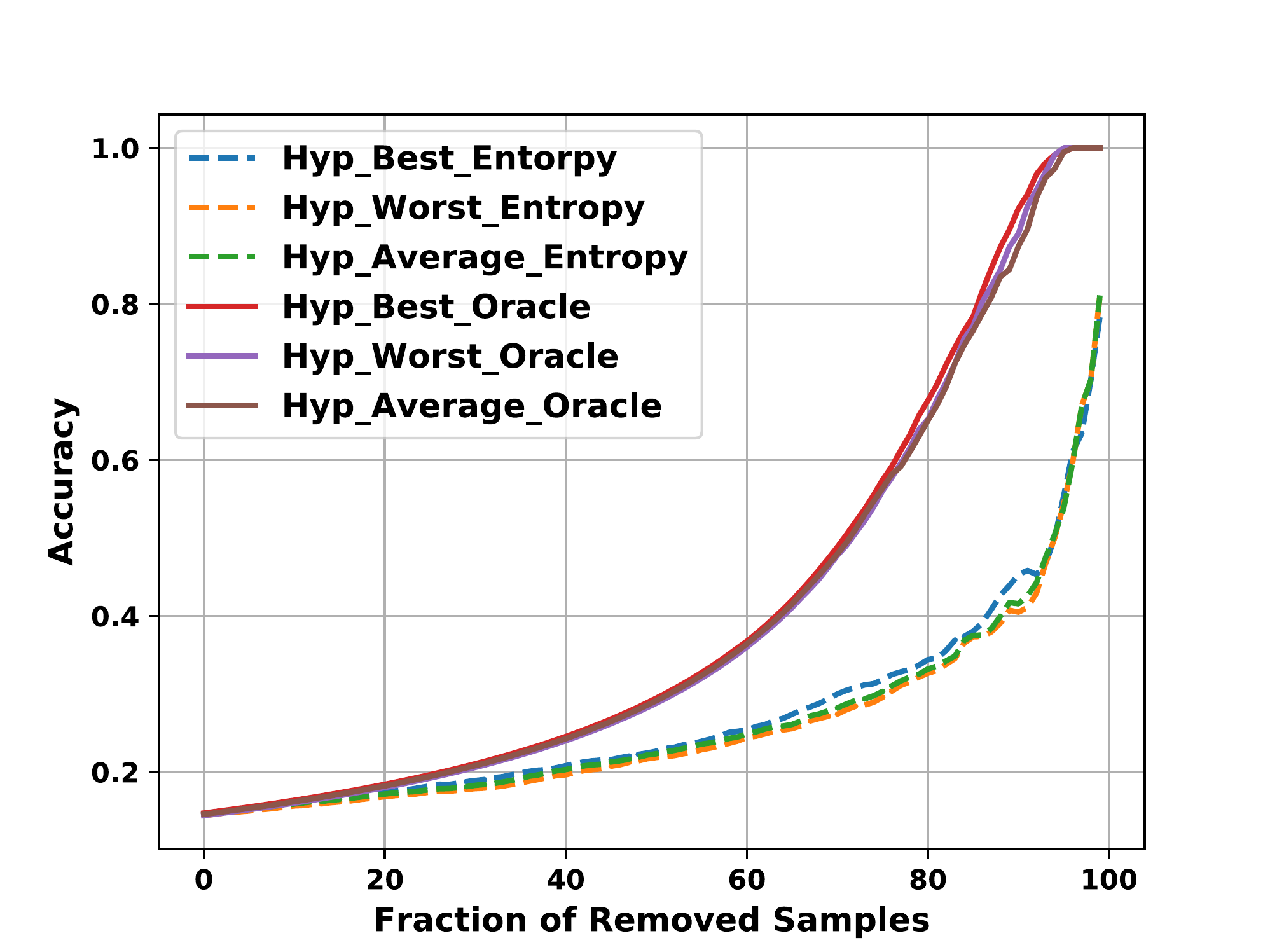}} 
\subfloat[EpicKitchen-Action\label{fig_sparsification:d}]{\includegraphics[width = 1.8in]{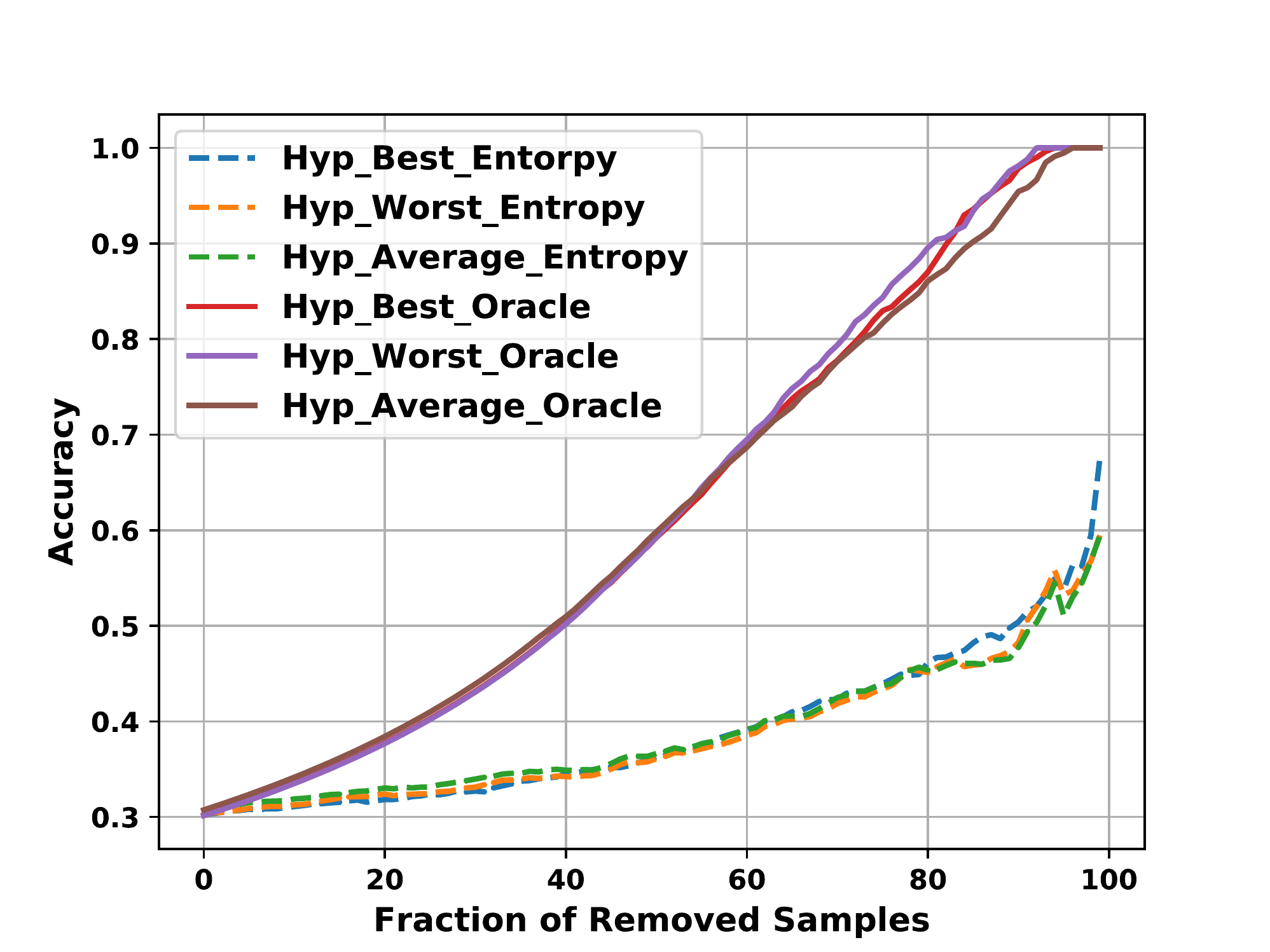}} 
       \caption{Sparsification plot of proposed method for the Breakfast ($a$ and $b$) and Epic-Kitchen ($c$ and $d$) datasets.The plot shows the accuracy of future activity prediction for each fraction of samples having the highest uncertainties removed. The oracle sparsification shows the upper bound by removing each fraction of samples ranked by the cross-entropy loss between the prediction and the ground-truth. The big difference to its oracle can be explained by the relatively big errors inherent in future prediction task.}
 \label{fig:class_sparse}
\end{figure}

\begin{figure}
\subfloat[Breakfast-Object\label{fig_sparsificationB:a}]{\includegraphics[width = 1.76in]{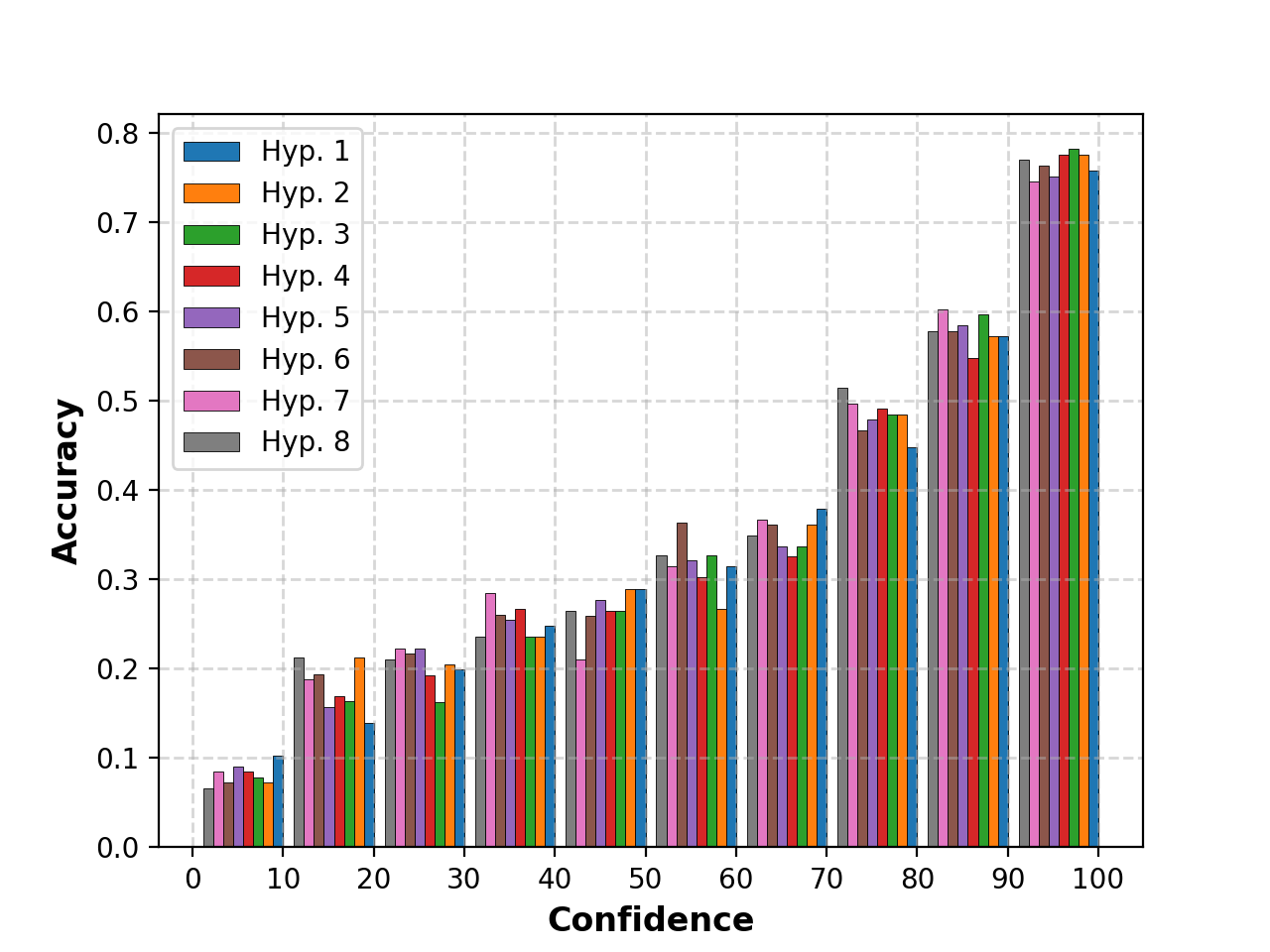}} 
\subfloat[Breakfast-Action\label{fig_sparsificationB:b}]{\includegraphics[width = 1.76in]{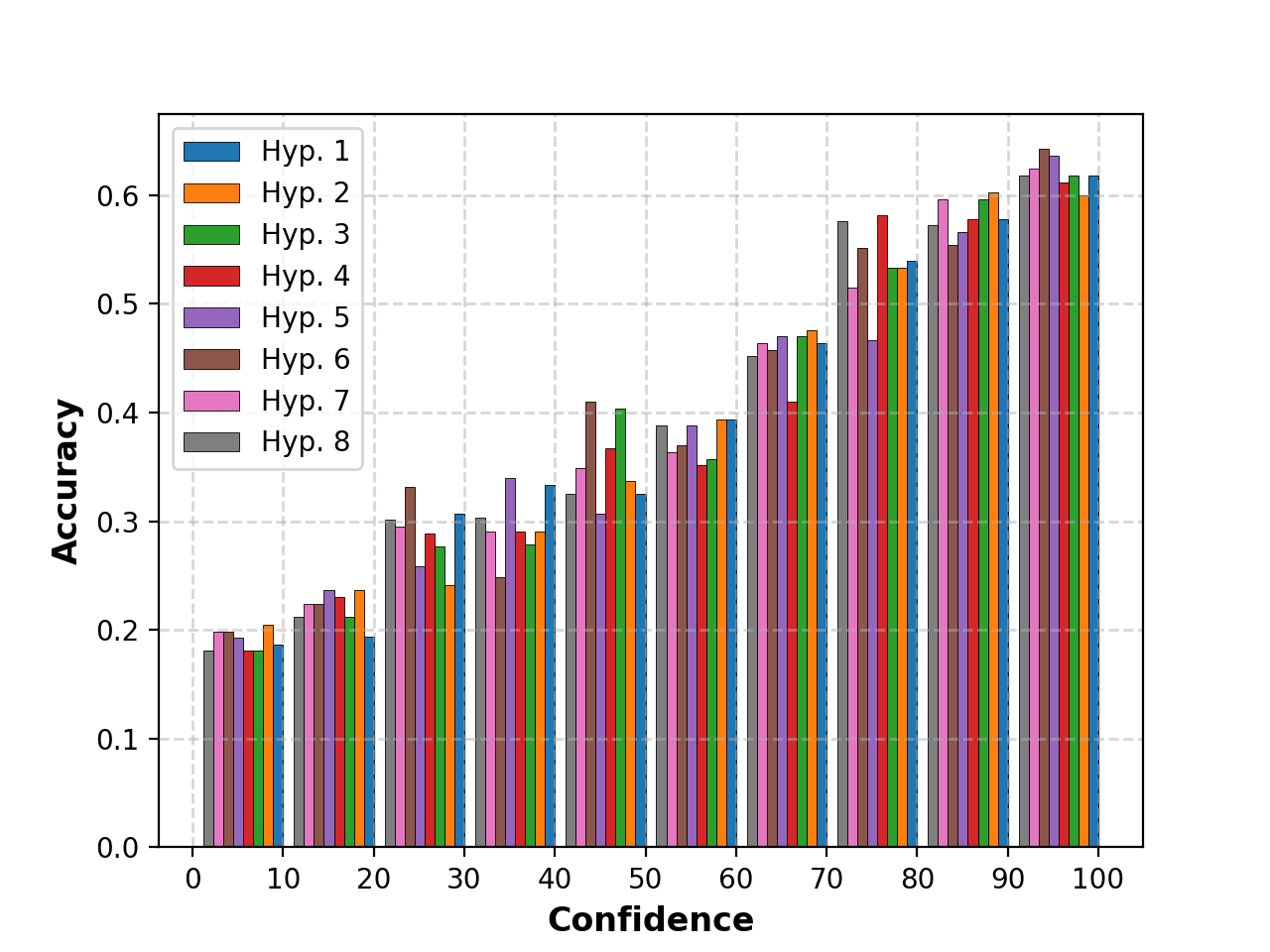}} \\ 
\subfloat[EpicKitchen-Object\label{fig_sparsificationB:c}]{\includegraphics[width = 1.76in]{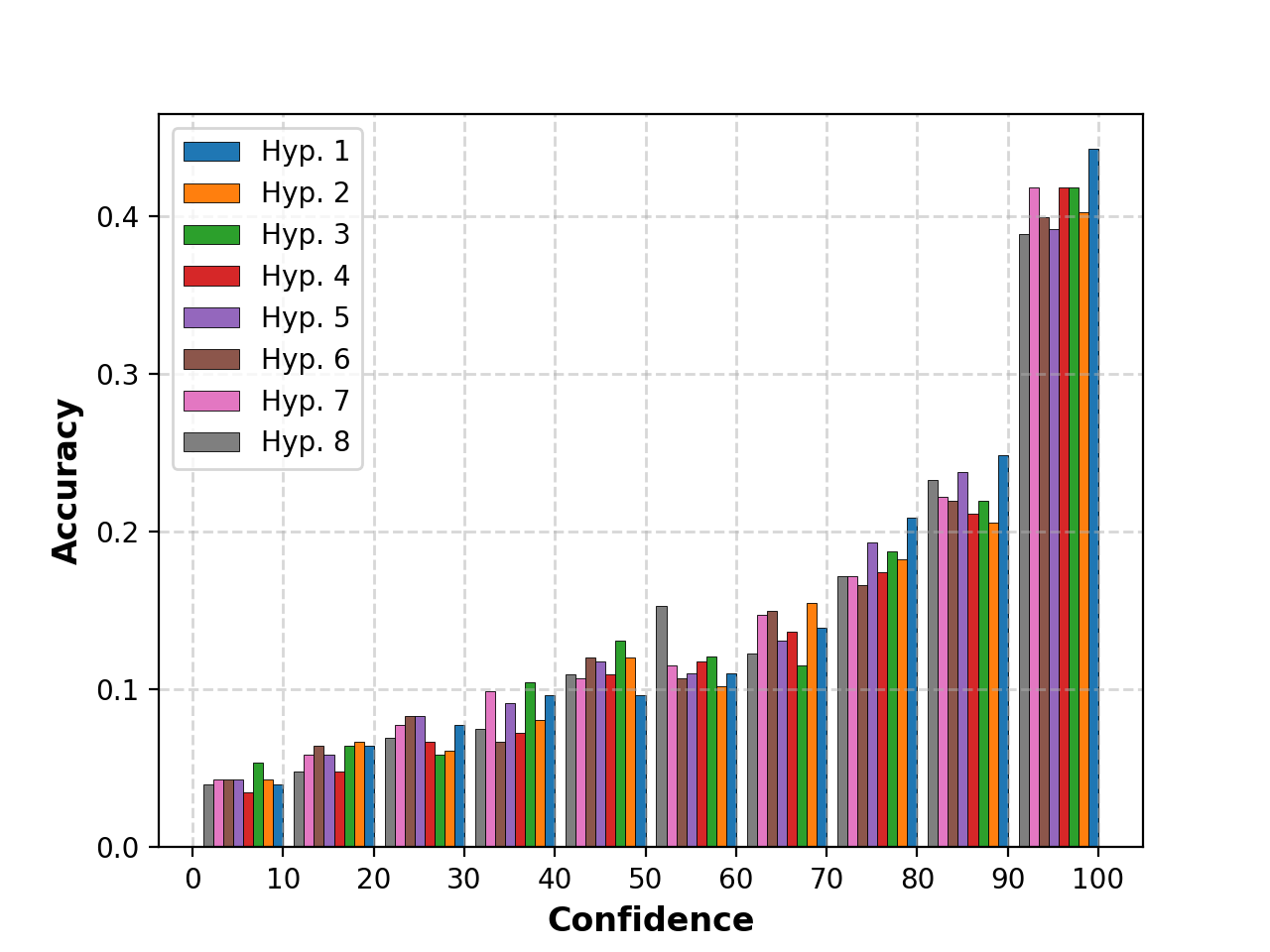}} 
\subfloat[EpicKitchen-Action\label{fig_sparsificationB:d}]{\includegraphics[width = 1.76in]{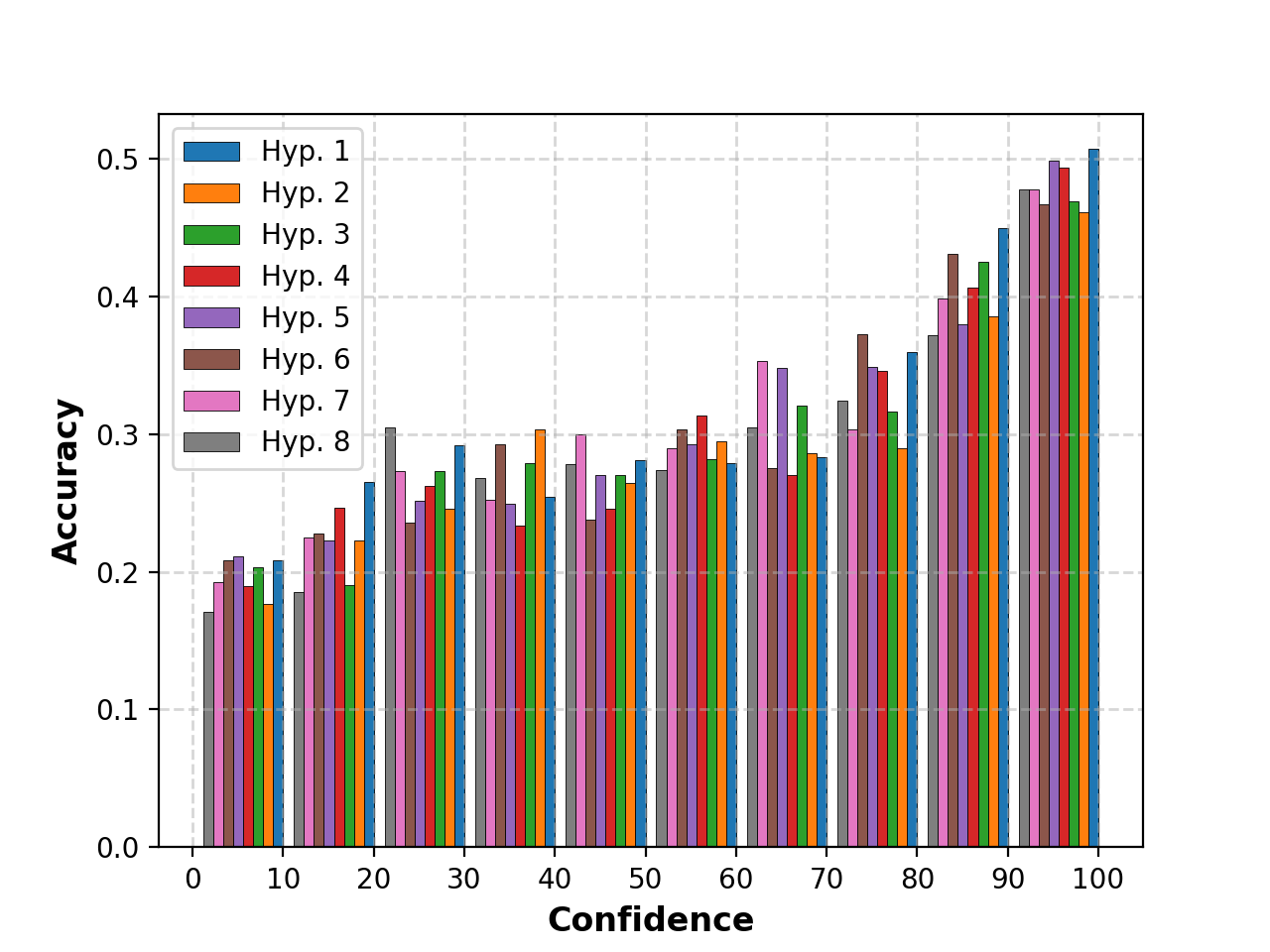}} 
       \caption{Reliability diagrams for Breakfast (a and b) and Epic-Kitchens (c and d) datasets. Diagonal increase in accuracy suggests that our uncertainties are well calibrated. By increasing the confidence threshold, accuracy increases consistently.  }
       \label{fig:binning_class}
\end{figure}

 \begin{figure}[htp]
\centering
\includegraphics[trim=0mm 0mm 0mm 0mm,clip,width=0.48\textwidth]{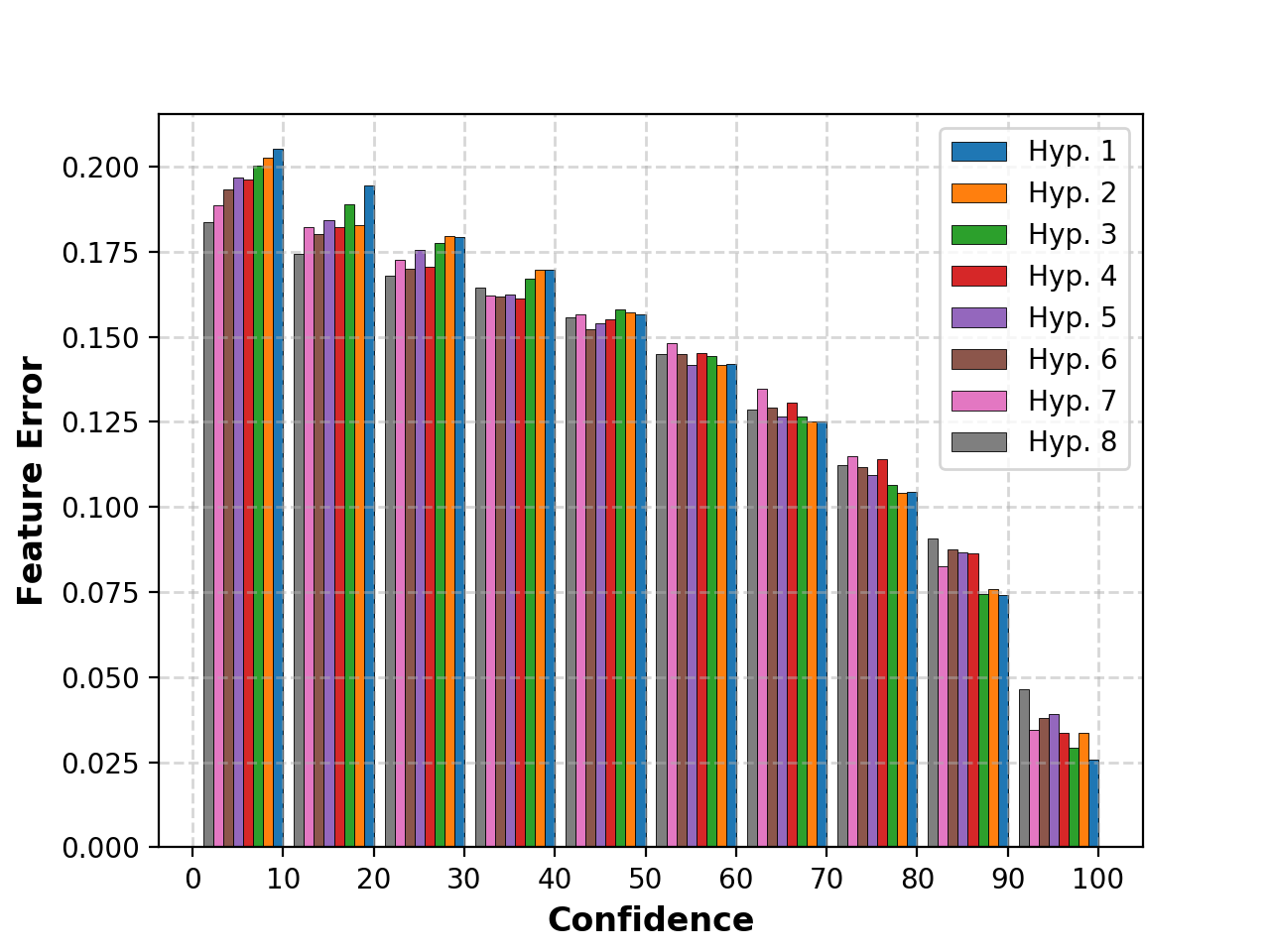}
\caption{ Reliability diagram for our feature uncertainties on Breakfast dataset. Diagonal decrease suggests that our uncertainties are well calibrated and potentially useful. }
\label{fig:bin}
\end{figure}

\begin{figure}
\subfloat[Breakfast\label{fig_sparsification:a}]{\includegraphics[width = 3.46in]{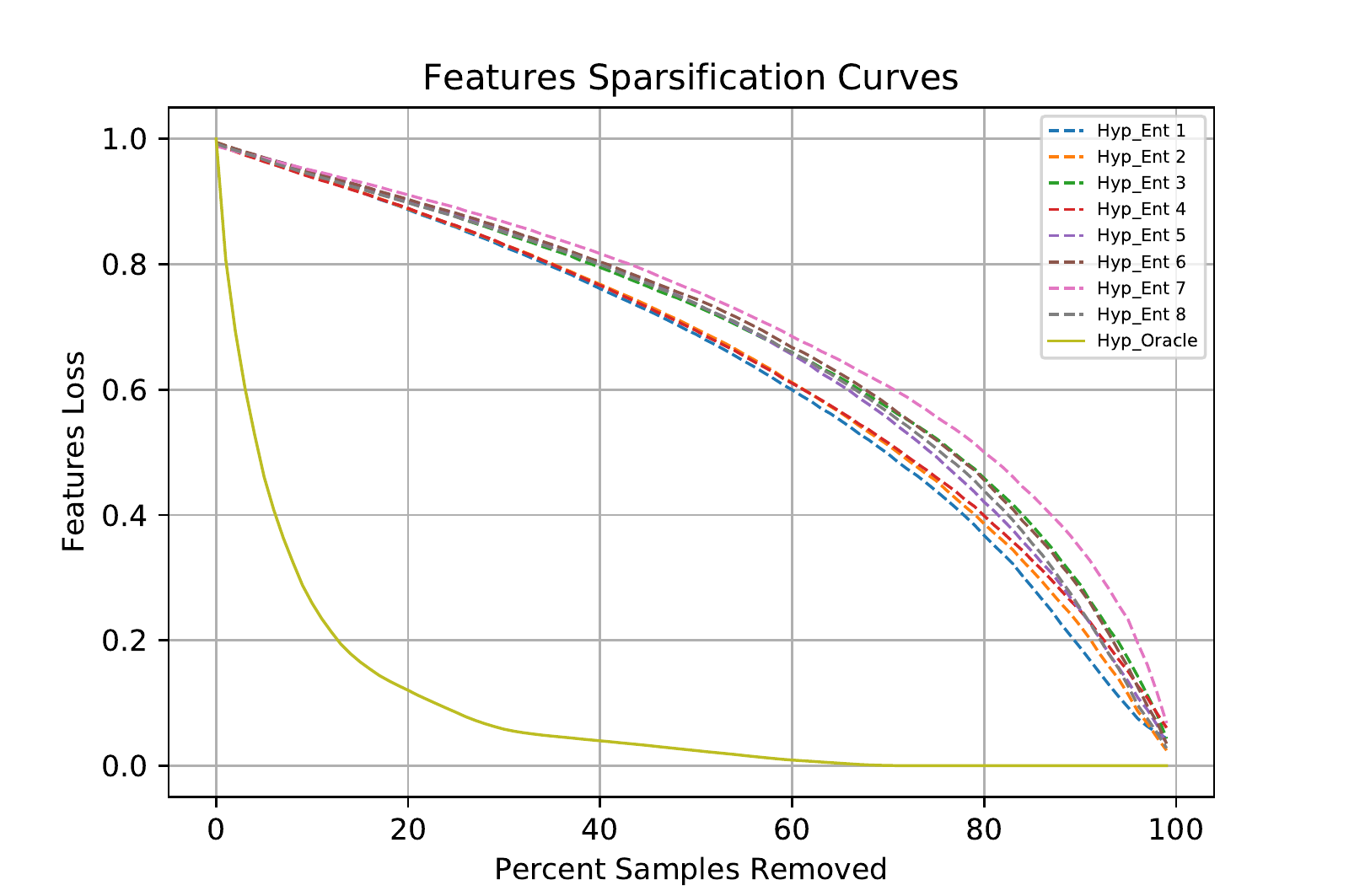}} \\
\subfloat[Epic-Kitchens\label{fig_sparsificationB:b}]{\includegraphics[width = 3.46in]{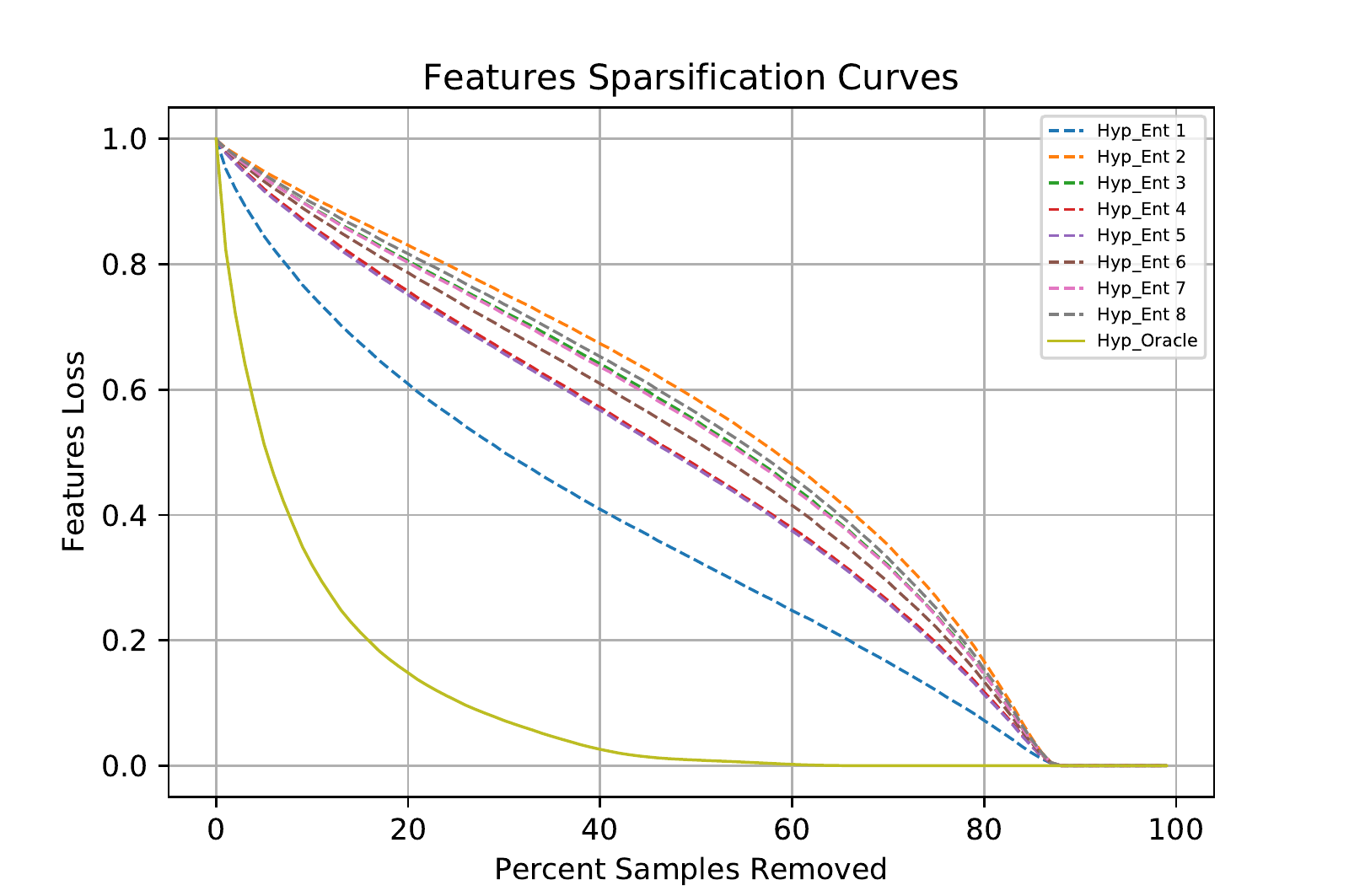}} 

\caption{Sparsification plot for feature reconstruction uncertainties for Breakfast (a) and Epic-Kitchens (b). The big difference to its oracle can be explained by the relatively big errors inherent in future prediction task.}
\label{fig:sparsification_features}
\end{figure}

  \begin{figure*}[t]
 \centering
 \includegraphics[trim=0mm 0mm 0mm 0mm,clip,width=1\textwidth]{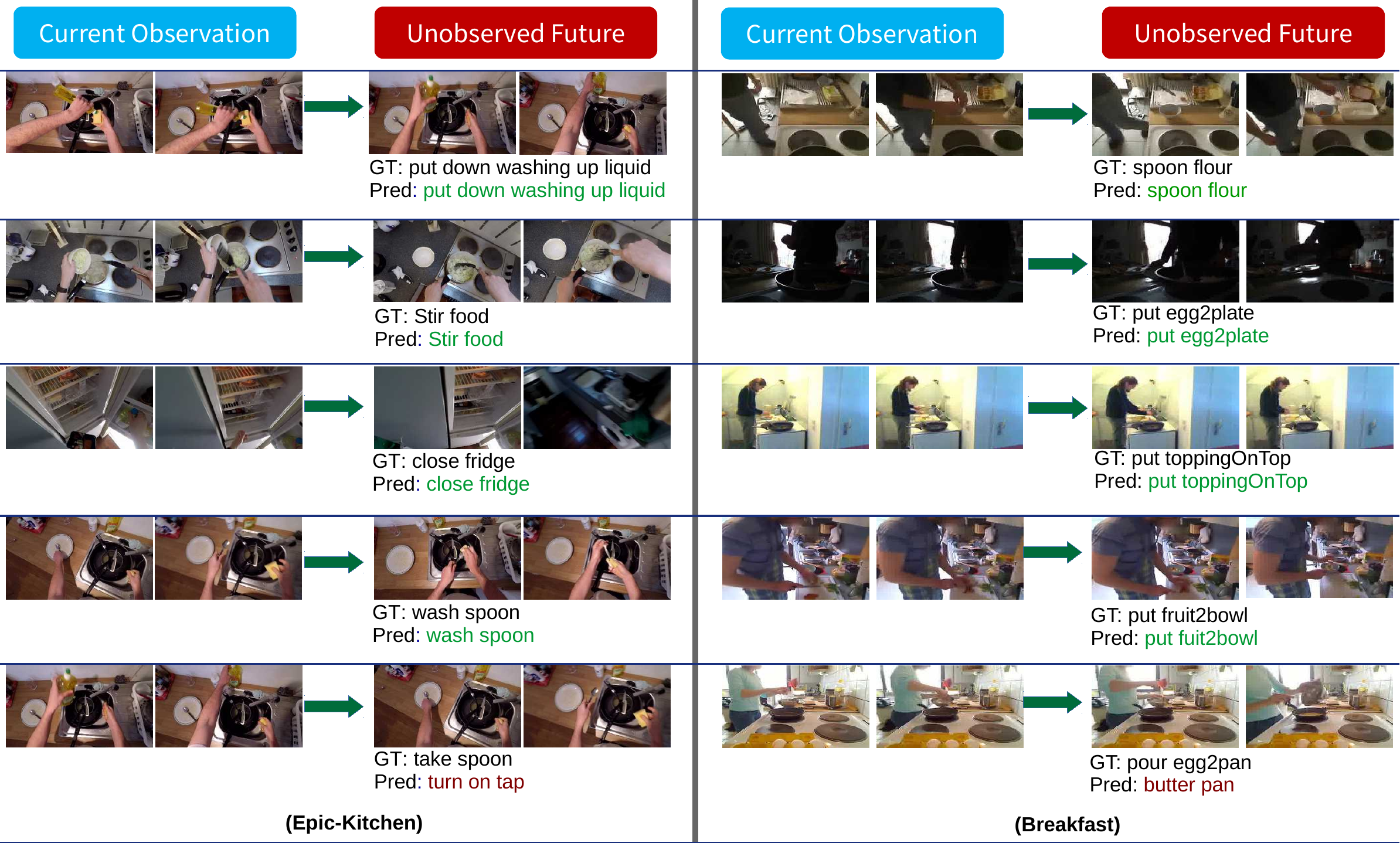}
 \caption{  Qualitative examples of future prediction on Epic-kitchens and Breakfast datasets. For each example, current observation and future observation are provided. Last row shows the failure examples of future prediction.  }
 \label{fig:qualitative_results2}
 \end{figure*}

\section{Uncertainty evaluation}

For evaluating the quality of the uncertainty predictions we use reliability diagrams~\cite{calibration} and sparsification plots~\cite{opticalFlowPAMI2012,probFlow,Kondermann2008,bootstrapOpticalFlow, mhp}. A reliability diagram plots the expected quality as a function of uncertainty. If the model is well-calibrated this plot should draw a diagonal decrease. A sparsification plot shows the quality gain over the course of removing the samples with the highest uncertainties gradually. In the best case, samples would be removed using the ground truth error and this can serve as the ground-truth (oracle) for the sparsification plot.

In Figure~\ref{fig:class_sparse} we show the sparsification plots of the best, the worst, and the average hypothesis for classification of actions and objects for Breakfast (first row) and EpicKitchens (second row) datasets. Plots tends to consistently increase as the uncertainties removed.  In order to assess the quality of these plots we also provide the \textit{Oracle Plot}. The oracle is simply repeating the sparsification with the true error instead of the uncertainties to get the upper bound for the uncertainties. Ideally the closer the sparsification plot is to its oracle is the better. One possible reason for the relatively bigger distance in our plots can be that activity prediction has still not reached its saturation ($70\%$ error) while image classification has (typically $<5\%$ error).

In Figure~\ref{fig:binning_class} we report the reliability diagram per hypothesis on both Breakfast and Epic-Kitchens datasets for future action/object classification. The diagonal increase suggests that our uncertainties are well calibrated. Accuracy tends to consistently increase as the confidence threshold of removed samples increase.

In Figure~\ref{fig:bin} we report the reliability diagram per hypothesis on Breakfast dataset for feature reconstruction. For Epic-Kitchens diagram, see the Figure 7 of the main paper. The diagonal decrease suggests that our uncertainties are useful as also supported by our captioning results.

In Figure~\ref{fig:sparsification_features} we show the sparsification plots per hypothesis for feature reconstruction for both datasets. Error tends to decrease as the high uncertain features removed. However, as in the classification case there is a big difference to its oracle due to the difficulty in future prediction. When the predictions do not generalize, uncertainties also do not.


 \section{Qualitative results for the future prediction}
 Figure~\ref{fig:qualitative_results2} shows representative results of our method. We input the current observation to the model and get the future captions.
 
 Epic-kitchens dataset is ego-centeric and camera mounted on the person's head. Therefore, future prediction on this dataset is more challenging due to the quick changes in view point and motion blurr. In the Breakfast dataset, the camera location is fixed throughout the video recording.
 
 In the last row of Figure~\ref{fig:qualitative_results2} for the Epic-Kitchens, it is not known that tap is on or off and for the Breakfast, pan being buttered is not known. This implies that providing longer history of previously performed actions would decrease the ambiguity of future prediction.
 
 For instance, in the last row for the Epic-Kitchens the observation is "put down washing liquid" and prediction is "turn on tap" while ground-truth is "take spoon".
